\pdfoutput=1

\documentclass[11pt]{article}
\usepackage[table]{xcolor}

\usepackage[preprint]{acl}

\usepackage[T1]{fontenc}

\usepackage[utf8]{inputenc}

\usepackage{microtype}

\usepackage{inconsolata}

\usepackage{graphicx}

\usepackage{xspace}
\usepackage{times}
\usepackage{booktabs}
\usepackage{latexsym}
\usepackage{graphicx}
\usepackage{multirow}
\usepackage{multicol}
\usepackage{hyperref}
\usepackage{listings}
\usepackage{amsmath}
\usepackage[most]{tcolorbox}
\usepackage{amssymb}
\usepackage{pifont}
\usepackage{textcomp}
\usepackage{enumitem}
\usepackage{subcaption}

\usepackage{pgfplots}
\pgfplotsset{width=10cm,compat=1.16}

\usepackage{CJKutf8}

\usepackage{listings}

\usepackage[capitalize]{cleveref}

\crefname{section}{Sec.}{Secs.}
\Crefname{section}{Section}{Sections}
\Crefname{table}{Table}{Tables}
\crefname{table}{Tab.}{Tabs.}
\crefname{figure}{Fig.}{Figs.}
\Crefname{figure}{Figure}{Figures}
\crefname{appendix}{Appx.}{Appxs.}
\Crefname{appendix}{Appendix}{Appendixs}
\crefname{subsection}{Sec.}{Secs.}

\def\etc{\emph{etc.}}

\newcommand{\tablestyle}[2]{\setlength{\tabcolsep}{#1}\renewcommand{\arraystretch}{#2}\centering\footnotesize}
\newlength\savewidth

\newcommand\blfootnote[1]{\begingroup\renewcommand\thefootnote{}\footnote{#1}\addtocounter{footnote}{-1}\endgroup}

\newcommand\bluecolor[1]{\cellcolor{gray!40!blue!30}{#1}}
\newcommand\greencolor[1]{\cellcolor{gray!40!green!30}{#1}}

\newcommand\purplecolor[1]{\cellcolor{gray!20!purple!30}{#1}}

\title{MathBench: Evaluating the Theory and Application Proficiency\\of LLMs with a Hierarchical Mathematics Benchmark}

\author{Hongwei Liu$^{1}$ \quad Zilong Zheng$^{1,3}$ \quad Yuxuan Qiao$^{1,4}$ \quad Haodong Duan$^{1}$ \quad Zhiwei Fei$^{1,4}$ \\
\textbf{ \quad Fengzhe Zhou$^{1}$ \quad Wenwei Zhang$^{1}$ \quad Songyang Zhang$^{1,\dagger}$ \quad Dahua Lin$^{1,2}$ \quad Kai Chen$^{1,\dagger} $} \\
 $^{1}$Shanghai AI Laboratory\quad $^{2}$The Chinese University of Hong Kong\\
 $^{3}$Beihang University \quad $^{4}$Nanjing University\\
}

\begin{document}
\maketitle
\blfootnote{\noindent $^{\dagger}$ Corresponding authors. 
}
\begin{abstract}

Recent advancements in large language models (LLMs) have showcased significant improvements in mathematics. 
However, traditional math benchmarks like GSM8k offer a unidimensional perspective, falling short in providing a holistic assessment of the LLMs' math capabilities. 
To address this gap, we introduce MathBench, a new benchmark that rigorously assesses the mathematical capabilities of large language models. 
MathBench spans a wide range of mathematical disciplines, offering a detailed evaluation of both theoretical understanding and practical problem-solving skills. 
The benchmark progresses through five distinct stages, from basic arithmetic to college mathematics, and is structured to evaluate models at various depths of knowledge. 
Each stage includes theoretical questions and application problems, allowing us to measure a model's mathematical proficiency and its ability to apply concepts in practical scenarios. 
MathBench aims to enhance the evaluation of LLMs' mathematical abilities, providing a nuanced view of their knowledge understanding levels and problem solving skills in a bilingual context. 
The project is released at \url{https://github.com/open-compass/MathBench}.

\end{abstract}
\section{Introduction}

Mathematical reasoning and problem-solving represent pivotal facets of human intelligence and have captivated the interest of artificial intelligence (AI) research for decades. 
The capability of machines to grasp, interpret, and address mathematical challenges not only serves as a benchmark for their cognitive prowess but also fulfills a critical role in their deployment across various sectors. 

\begin{figure}[!ht]
    \centering
    \includegraphics[width=1\linewidth]{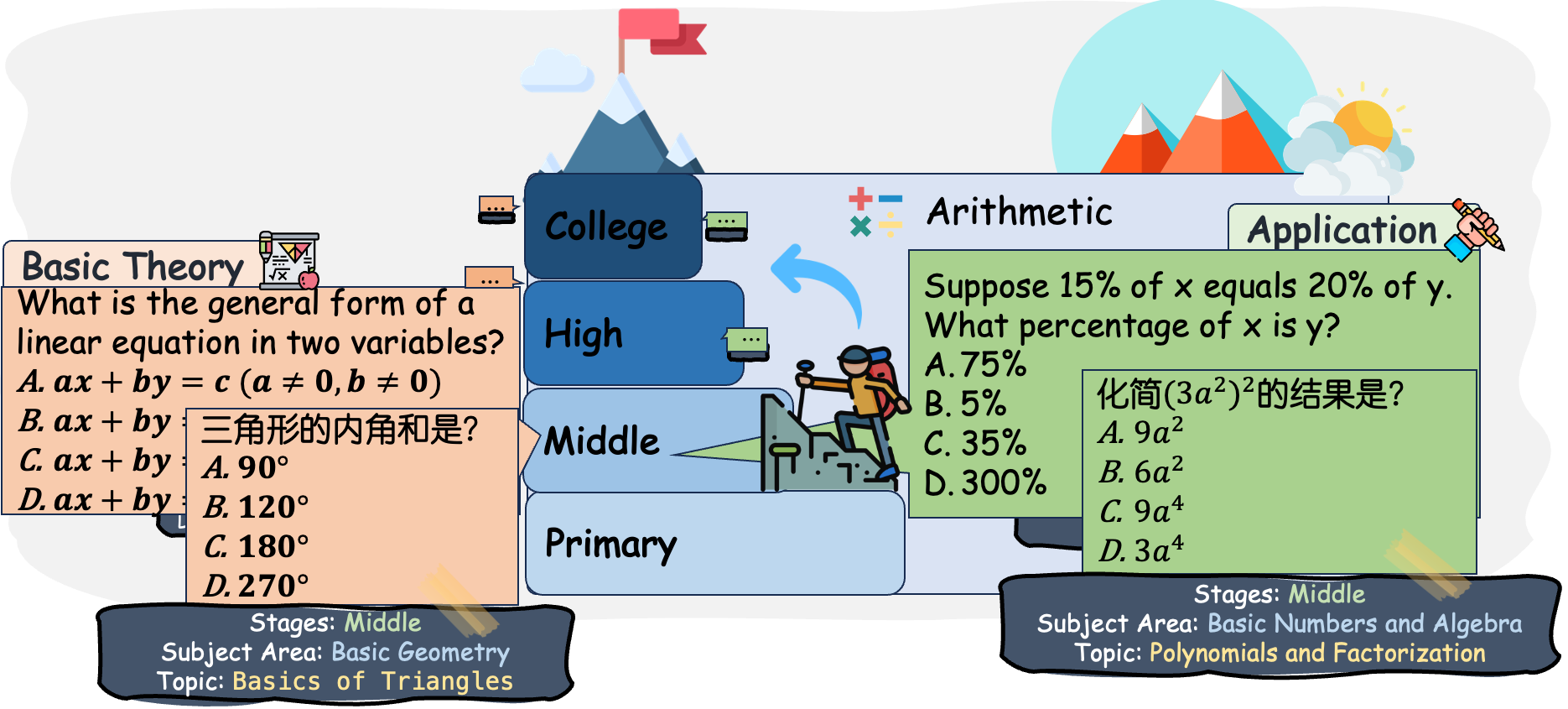}
    \caption{\textbf{MathBench Overview. } 
    MathBench comprises multiple stages of progressively increasing challenges. 
    Each stage encompasses bilingual theoretical and application-oriented questions, with each question precisely tagged with a three-level label to indicate its fine-grained knowledge point.}
    \label{fig:comic_stages}
\end{figure}


The advent of modern Large Language Models (LLMs) such as OpenAI's ChatGPT and GPT-4 \citep{achiam2023gpt} has marked a significant milestone,
showcasing an unparalleled ability to generate text that mirrors human-like discourse and to unravel intricate mathematical conundrums \citep{mathematic_survey}.

Despite these advancements, the evaluation of LLMs' mathematical capabilities remains hampered by some inherent limitations of existing benchmarks (GSM8k \citep{gsm8k}, MathQA\citep{MathQA}, \etc). 
These resources predominantly offer a singular perspective on problem-solving abilities and lack comprehensive difficulty grading. 
Math \citep{math} attempted to classify high-school math competition problems into varying levels of complexity based on annotators' subjective evaluations, offering an incomplete picture of mathematical proficiency. 
Such datasets, while valuable, fall short in encapsulating the full spectrum of mathematical knowledge and
overlook the importance of fundamental theory understanding, which is essential for tackling application problems \citep{draw}. 
Those limitations make it difficult to conduct a comprehensive evaluation of LLMs' math capability (both theory and application) across different levels and disciplines and under a multilingual context.


\begin{figure*}[!ht]
    \centering
    \includegraphics[width=\linewidth]{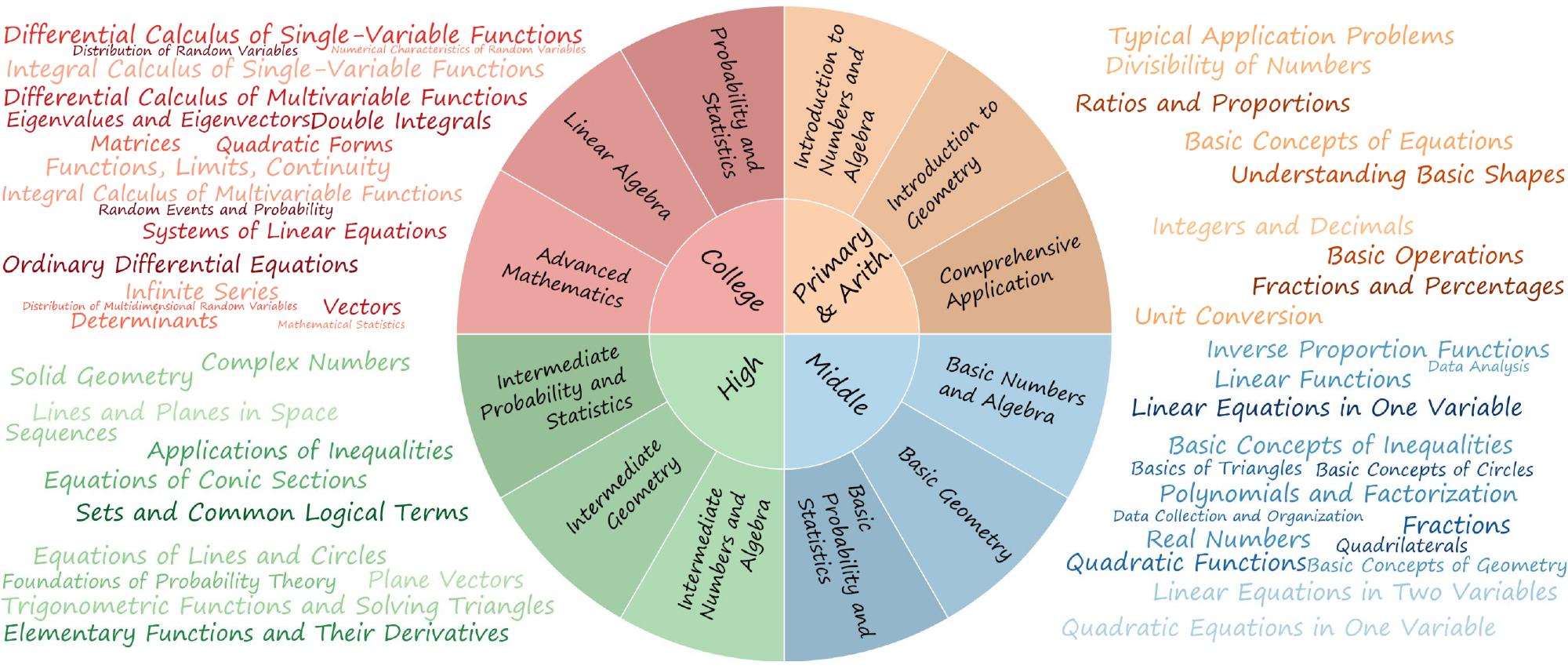}
    \caption{\textbf{Framework of MathBench}, We first categorize the mathematical content into four main educational stages and one basic arithmetic stage. 
    Then, we extend from these to fill in two more fine-grained levels of knowledge points, 
    forming the final MathBench framework. }
    \label{fig:mathbench_overall}
\end{figure*}

In response to these challenges, we construct \textit{MathBench}, a novel and comprehensive multilingual benchmark meticulously created to evaluate the mathematical capabilities of LLMs across a diverse range of difficulties,  from basic arithmetic to challenging college-level mathematics. 
\textit{MathBench} sets itself apart with a unique five-stage taxonomy, mapped to the educational trajectory from primary school through to college. 
This mechanism is designed to assess LLMs' mathematical understanding in breadth and depth.
The benchmark incorporates carefully curated questions that cover basic theory knowledge and practical applications. 
This dual focus enables \textit{MathBench} to probe and interpret the models' capabilities from a foundational standpoint. 
Additionally, \textit{MathBench} supports bilingual evaluation in both Chinese and English, which facilitates a more nuanced and comprehensive assessment of LLMs' math capabilities, offering a realistic reflection of the global landscape of mathematical knowledge.

In this paper, we detail the methodology behind the creation of MathBench, including the hierarchical knowledge system that underpins the dataset, the data collection process, and the criteria for question selection. 
We hope that MathBench can serve as a valuable resource for researchers and developers seeking to advance the mathematical abilities of LLMs and to understand the limitations of existing models in solving diverse and complex mathematical problems.



\textit{MathBench} features the following contributions:

\begin{enumerate}[label={\bf {{$\bullet$}}},leftmargin=*,topsep=0.5ex,itemsep=-0.5ex,partopsep=0.75ex,parsep=0.75ex,partopsep=0pt,wide,labelindent=0pt]
    \item We introduce \textit{MathBench}, a comprehensive dataset that features a five-level difficulty mechanism with a hierarchical knowledge system. 
    \item MathBench includes a wide variety of question types, from  fundamental mathematical concepts to practical application in real-world scenarios.
    \item We conduct extensive experiments on MathBench across different models to identify bottlenecks in current LLMs. The provided discussion and analysis are expected to offer new avenues for improving their mathematical capabilities.
\end{enumerate}


\section{Methodology}

MathBench features a well-crafted difficulty hierarchy and an emphasis on evaluating the theoretical knowledge understanding of LLMs. 
\Cref{sec:knowledge_system} presents the tiered levels and the corresponding knowledge foundations, explaining the ability taxonomy and design rationale. 
\Cref{sec:data_collection} details the collection process and statistics of MathBench.

\subsection{The Hierarchical Knowledge System}
\label{sec:knowledge_system}

\begin{table}[!ht]
\centering
\caption{\textbf{Overview of Datasets Included in MathBench}. MCQ stands for Multi-Choice Question.}
\label{tab:datasets_overview}
\resizebox{\columnwidth}{!}{
\tablestyle{6pt}{1.2}
\begin{tabular}{lll} 
\toprule
\textbf{Name} & \textbf{Dataset Type} & \textbf{Question Type} \\ 
\midrule
GSM-X-CN & Self-Collected & Open-ended QA \\
GSM-X-Plus & Self-Collected & Open-ended QA \\
CEVAL-Math & Open Source & MCQ \\
MMLU-College-Math & Open Source & MCQ \\
Math401 & Open Source & MCQ \\
Hungarian-Math-MCQ & Self-Collected & MCQ \\
AMC-8 \& 12 & Self-Collected & MCQ \\
SAT & Self-Collected & MCQ \\
Gaokao & Self-Collected & MCQ \\
Zhongkao & Self-Collected & MCQ \\
Kaoyan & Self-Collected & MCQ \\
SciBench & Open Source & MCQ \\
Arithmetic-HG & Open Source & Open-ended QA \\
Theory-Knowledge-Primary & Self-Collected & MCQ \\
Theory-Knowledge-Middle & Self-Collected & MCQ \\
Theory-Knowledge-High & Self-Collected & MCQ \\
Theory-Knowledge-College & Self-Collected & MCQ \\
\bottomrule
\end{tabular}}
\vspace{-5mm}
\end{table}

In MathBench, we define a knowledge framework with five main stages and three levels in order to obtain fine-grained evaluation results.
Among five stages, four stages are mapped to the \textbf{four main education stages}: \textit{Primary, Middle, High, and College}, 
while the other stage is named \textit{Arithmetic}, serving as the foundation of the remaining four stages.\footnote{The `Arithmetic' stage evaluates the ability to perform four basic math operations: add, subtract, multiply, divide. }
Each \textbf{Stage} in MathBench is associated with two fine-grained knowledge levels: \textbf{Subject Area} and \textbf{Topic}, accordingly.
As shown in \Cref{fig:mathbench_overall}, we extend MathBench from the basic stages to a comprehensive range of mathematical concepts and problem-solving skills. 
Such taxonomy is designed to capture the depth and breadth of mathematical knowledge, from foundational arithmetic to complex, abstract college-level concepts.


\textbf{Subject Areas} include major mathematical disciplines such as Algebra, Geometry, Trigonometry, Calculus, Statistics, Probability, \etc.  
This categorization allows for a wide range of questions, facilitating an organized approach to covering the diverse areas of mathematics.
Within each subject area, we further refine the classification into specific \textbf{Topics}. 
For example, under Algebra, topics might include Linear Equations, Quadratic Equations, Polynomials, and Functions. 
The Topic-level granularity ensures that the dataset can provide detailed insights into a model's understanding and proficiency in specific areas of mathematics.

In MathBench, each question is tagged with metadata indicating its stage (Primary, Middle, High, College, or Arithmetic), subject area, and topic. 
Such tags enable a fine-grained analysis of models' performance across different areas of mathematics and allow researchers to identify specific strengths and weaknesses in mathematical understanding.

Moreover, the inclusion of the Arithmetic stage emphasizes the importance of mastering basic math operations, 
which is the foundation of all subsequent mathematical learning and problem-solving. 


\subsection{Data Collection and Statistics}
\label{sec:data_collection}

With the pre-defined knowledge framework, we primarily collect questions from two perspectives: 
\textit{(a). theoretical knowledge questions}, to test the model's grasp of basic formulas, theories, and their corollaries, which are the foundation for solving mathematical problems; 
\textit{(b). practical application questions}, which often require a good understanding of the fundamental theories, reflecting the ability to apply these theories in practice.

\paragraph{Question Format Definition. }

During the evaluation, 
some models struggle with open-ended questions and fail to follow instructions and provide plain and concise answers.
Therefore, we reformulate questions that could have complex answers\footnote{All theoretical knowledge questions and practical application questions from middle school to college level} into the multiple-choice format, typically with four options. 
During collection and annotation, we ensure the uniqueness of the correct answer and the high confusion-level of distractive options. 

\paragraph{Theoretical Knowledge Questions. }
\label{par: Theoretical Knowledge Questions}
For theoretical knowledge questions, we collect the definition and detailed corollaries of knowledge points topic by topic from the math textbooks and the Internet. 
We then transform them to multi-choice questions with high-quality annotations.  

\paragraph{Practical Application Questions. }
On selecting the practical application questions, we primarily consider the following aspects: 
1. The question needs to match the corresponding education level;
2. The questions should comprehensively cover the previously defined knowledge taxonomy;
3. The questions should be well-formulated so that LLMs can answer them properly.
We primarily focus on stage-based educational exams or competitions. 
Those questions are comprehensive and representative, offering a certain degree of difficulty gradient, 
such as ZhongKao, GaoKao in Chinese Math and AMC, SAT in English math. 
Additionally, we incorporate open-source questions to enhance the diversity and breadth of the questions. 
We list the sources of questions in MathBench in \Cref{tab:datasets_overview}.

\paragraph{Quality Screening. }
To enhance the quality of the MathBench dataset, we implement a semi-automated question filtering process to mitigate issues such as intrinsic question errors and alignment with educational stages utilizing GPT-4, details presented in \Cref{app: quality_screening}.

\paragraph{Dataset Summary. }
We curate 3709 questions for the final MathBench, including both Chinese and English languages across five stages with three-level knowledge taxonomy. 
This compendium is divided into two distinct sections: MathBench-T, which consists of 2,209 theoretical questions, and MathBench-A, comprising 1,500 questions focused on practical applications. Each question has been subjected to a rigorous semi-automated vetting process.  Detailed statistics can be found in the Appendix \ref{app:data_Statistics}.


\section{Experiments and Analysis}

\subsection{Configuration}

\paragraph{Evaluation Protocols. }
We employ CircularEval (CE)~\citep{mmbench} and Perplexity (PPL) as our principal evaluation methodology for Chat and Base models respectively.
CE systematically assesses an $N$-option multi-choice question by evaluating it $N$ times, each time permuting the order of the options. 


To maintain consistency in evaluations, we standardized the maximum output length to 2048 tokens and employed a greedy decoding strategy for all Large Language Models (LLMs). For open-ended questions, we utilized a few-shot CoT setting, whereas for multiple-choice questions on Chat models, we implemented a zero-shot CoT approach. In the case of Base models during PPL evaluation, a few-shot setting was adopted. We used OpenCompass~\citep{2023opencompass} as the evaluation framework for our assessments.

\paragraph{Evaluated Models. }
Our evaluation encompasses both closed-source commercial LLMs and open-source LLMs, covering more than $20$ models. 
Based on MathBench, we deliver a thorough evaluation of the capabilities of current LLMs.
We list all evaluated LLMs below:


\noindent$\bullet$ Closed-source models: GPT-3.5 and GPT-4\footnote{GPT-4 version: \texttt{gpt-4-0125-preview and GPT-4o(GPT-4o-2024-05-13)}; GPT-3.5 version: \texttt{gpt-3.5-turbo-0125}}from Openai, Qwen-Max\footnote{\url{https://help.aliyun.com/zh/dashscope/create-a-chat-foundation-model?spm=a2c4g.11186623.0.0.581c64d16b7Azw}}, DeepSeek-V2-API\footnote{\url{https://platform.deepseek.com/api-docs}}, GLM4\footnote{\url{https://open.bigmodel.cn/dev/howuse/glm4}} and Anthropic Claude-3-Opus\footnote{\url{https://www.anthropic.com/news/claude-3-family}}.

\noindent$\bullet$ OpenSource LLMs: We evaluate a wide spectrum of LLMs, including Llama3\cite{touvron2023llama2}, Qwen \cite{bai2023qwen}, InternLM2 \cite{team2023internlm}, Yi \footnote{\url{https://github.com/01-ai/Yi}}, Baichuan2~\citep{yang2023baichuan},  DeepSeek\cite{deepseek}, Mixtral\cite{jiang2024mixtral}and ChatGLM3 \cite{zeng2022glm}.

\noindent$\bullet$ OpenSource Math LLMs: Llemma\cite{azerbayev2023llemma}, MetaMath-llemma\cite{MetaMath}, DeepSeek-Math\cite{deepseekmath}, MAmmoTH\cite{mammoth} and InternLM2-Math\citep{ying2024internlmmath}.

\subsection{Main Results}
We showcase the principal outcomes of MathBench in \Cref{tab: apllication_theory_chat_result}, detailing the application-oriented aspects in (\textit{MathBench-A}), and the theoretical components in (\textit{MathBench-T}).

\begin{table*}[!htb]
    \begin{minipage}{.52\linewidth}
      \centering
        \resizebox{\linewidth}{!}{
\begin{tabular}{lccccc|c}
\hline
\textbf{Models} & \textbf{Arith} & \textbf{Primary} & \textbf{Middle} & \textbf{High} & \textbf{College} & \textbf{Avg.} \\
\hline
\multicolumn{7}{c}{\purplecolor{\textit{\ding{72}Closed-source Models}}} \\
GPT-3.5-Turbo-0125 & 72.7 & 72.3 & 27.3 & 18.3 & 14.3 & 41.0 \\
GLM4 & 61.7 & 80.0 & 55.7 & 38.7 & 20.7 & 51.3 \\
GPT-4-0125-Preview & 76.0 & 82.3 & 59.0 & 41.3 & 35.3 & 58.8 \\
Qwen-Max-0428 & 72.3 & 86.3 & 65.0 & 45.0 & 27.3 & 59.2 \\
DeepSeek-V2-API & 82.7 & \textbf{89.3} & 59.0 & 39.3 & 29.3 & 59.9 \\
Claude-3-Opus & \textbf{85.7} & 85.0 & 58.0 & 42.7 & 43.7 & 63.0 \\
GPT-4o-2024-05-13 & 77.7 & 87.7 & \textbf{76.3} & \textbf{59.0} & \textbf{54.0} & \textbf{70.9} \\
\hline
\multicolumn{7}{c}{\bluecolor{\textit{$\heartsuit$Open-source Chat Models}}} \\
Yi-6B-Chat & 35.3 & 36.3 & 7.0 & 3.0 & 4.3 & 17.2 \\
ChatGLM3-6B & 38.0 & 41.0 & 13.7 & 5.3 & 1.7 & 19.9 \\
DeepSeek-7B-Chat & 48.3 & 47.7 & 8.7 & 4.3 & 2.7 & 22.3 \\
Qwen-7B-Chat & 50.7 & 50.7 & 22.0 & 9.3 & 6.0 & 27.7 \\
InternLM2-Chat-7B & 52.0 & 66.3 & \underline{30.0} & {13.7} & {8.7} & {34.1} \\
Llama-3-8B-Instruct & \underline{54.7} & \underline{71.0} & 25.0 & \underline{19.0} & \underline{14.0} & \underline{36.7} \\
\hline
Baichuan2-13B-Chat & 40.0 & 44.7 & 13.7 & 4.7 & 1.7 & 20.9 \\
Yi-34B-Chat & 50.7 & 62.0 & 23.0 & 14.7 & 7.7 & 31.6 \\
Qwen-14B-Chat & \underline{63.7} & 61.7 & \underline{39.0} & 21.0 & 12.0 & 39.5 \\
InternLM2-Chat-20B & 62.3 & \underline{72.7} & 37.7 & \underline{24.7} & \underline{13.0} & \underline{42.1} \\
\hline
DeepSeek-67B-Chat & 62.0 & 72.7 & 33.3 & 21.3 & 12.0 & 40.3 \\
Qwen-72B-Chat & \textbf{72.0} & 71.7 & 53.7 & 32.0 & 19.0 & 49.7 \\
Llama-3-70B-Instruct & 70.3 & \textbf{86.0} & 53.0 & 38.7 & \textbf{34.0} & 56.4 \\
Qwen1.5-110B-Chat & 70.3 & 82.3 & \textbf{64.0} & \textbf{47.3} & 28.0 & \textbf{58.4} \\
\hline
\multicolumn{7}{c}{\greencolor{\textit{$\triangle$Mathematical Models}}} \\
MammoTH-7B & 27.0 & 24.3 & 2.7 & 1.7 & 0.7 & 11.3 \\
MammoTH-13B & 35.0 & 43.0 & 5.0 & 4.7 & 5.0 & 18.5 \\
MammoTH-70B & 35.7 & 60.0 & 11.0 & 10.7 & 6.0 & 24.7 \\
Metamath-Llemma-7B & 51.7 & 51.0 & 8.3 & 8.3 & 5.0 & 24.9 \\
InternLM2-Chat-Math-7B & 53.7 & 67.0 & 41.3 & 18.3 & 8.0 & 37.7 \\
DeepSeek-Math-7B-Instruct & 61.0 & 74.0 & 30.3 & 24.7 & 14.3 & 40.9 \\
InternLM2-Chat-Math-20B & 58.7 & 70.0 & 43.7 & 24.7 & 12.7 & 41.9 \\
DeepSeek-Math-7B-RL & \underline{68.0} & \underline{83.3} & \underline{44.3} & \underline{33.0} & \underline{23.0} & \underline{50.3} \\
\hline
\end{tabular}
}
    \caption*{\small{MathBench-A.}}
    \end{minipage}%
    \hfill
    \begin{minipage}{.47\linewidth}
      \centering
        
        \resizebox{\linewidth}{!}{
\begin{tabular}{lcccc|c}
\hline
\textbf{Models} & \textbf{Primary} & \textbf{Middle} & \textbf{High} & \textbf{College} & \textbf{Avg.} \\
\hline
\multicolumn{6}{c}{\purplecolor{\textit{\ding{72}Closed-source Models}}} \\
GPT-3.5-Turbo-0125 & 70.1 & 56.7 & 47.3 & 52.5 & 56.7 \\
GLM4 & 88.6 & 79.5 & 63.7 & 60.6 & 73.1 \\
GPT-4-0125-Preview & 87.2 & 81.0 & 72.0 & 73.3 & 78.4 \\
Claude-3-Opus & 86.0 & 79.0 & 72.6 & 77.4 & 78.7 \\
DeepSeek-V2-API & 88.9 & 83.7 & 70.3 & 76.3 & 79.8 \\
Qwen-Max-0428 & 90.4 & 83.2 & 73.4 & 74.8 & 80.4 \\
GPT-4o-2024-05-13 & \textbf{92.2} & \textbf{88.3} & \textbf{82.0} & \textbf{85.6} & \textbf{87.0} \\
\hline
\multicolumn{6}{c}{\bluecolor{\textit{$\heartsuit$Open-source Chat Models}}} \\
DeepSeek-7B-Chat & 33.3 & 26.0 & 14.4 & 13.6 & 21.8 \\
ChatGLM3-6B & 41.6 & 32.4 & 20.2 & 12.0 & 26.6 \\
Yi-6B-Chat & 48.0 & 33.5 & 21.8 & 23.9 & 31.8 \\
Qwen-7B-Chat & 53.1 & 43.5 & 32.9 & 31.2 & 40.2 \\
Llama-3-8B-Instruct & 60.2 & 51.3 & 43.5 & 53.6 & 52.1 \\
InternLM2-Chat-7B & \underline{67.3} & \underline{55.8} & \underline{45.4} & \underline{42.7} & \underline{52.8} \\
\hline
Baichuan2-13B-Chat & 45.4 & 36.9 & 24.1 & 21.0 & 31.9 \\
InternLM2-Chat-20B & 64.5 & 56.2 & \underline{49.9} & 43.2 & 53.4 \\
Yi-34B-Chat & 70.9 & 57.0 & 43.6 & 46.8 & 54.6 \\
Qwen-14B-Chat & \underline{71.6} & \underline{64.0} & 49.7 & 49.4 & \underline{58.7} \\
\hline
DeepSeek-67B-Chat & 78.1 & 65.7 & 55.6 & 64.6 & 66.0 \\
Llama-3-70B-Instruct & 71.4 & 64.3 & 62.1 & 71.2 & 67.2 \\
Qwen-72B-Chat & 90.9 & 80.9 & 67.1 & 69.8 & 77.2 \\
Qwen-1.5-110B-Chat & \textbf{93.4} & \textbf{85.0} & \textbf{76.5} & \textbf{81.5} & \textbf{84.1} \\
\hline
\multicolumn{6}{c}{\greencolor{\textit{$\triangle$Mathematical Models}}} \\
MammoTH-7B & 11.6 & 9.1 & 8.4 & 6.3 & 8.8 \\
MammoTH-13B & 27.5 & 18.6 & 15.0 & 17.1 & 19.5 \\
MetaMath-Llemma-7B & 36.6 & 33.5 & 28.8 & 25.9 & 31.2 \\
MammoTH-70B & 58.1 & 47.1 & 39.3 & 44.6 & 47.3 \\
InternLM2-Chat-Math-7B & 65.6 & 60.2 & 51.7 & 46.5 & 56.0 \\
DeepSeek-Math-7B-Instruct & 73.3 & 58.4 & 49.3 & 50.3 & 57.8 \\
InternLM2-Chat-Math-20B & 73.2 & 70.5 & 60.6 & 53.0 & 64.3 \\
DeepSeek-Math-7B-RL & \underline{79.6} & \underline{72.0} & \underline{61.3} & \underline{68.7} & \underline{70.4} \\
\hline
\end{tabular}
        }
        \caption*{\small{MathBench-T.}}
    \end{minipage}
        \caption{\textbf{Overall Comparison of Models on MathBench-A \& T}. The \textit{Arithmetic} and \textit{Primary} stage for MathBench-T are combined because they share the same theory knowledge. Models are classified into three categories according to their purpose and origin. The model name in \textbf{bold} indicates the top performer among Open-source or Closed-source models, while an \underline{underline} signifies the leading model within a similar parameter size group.}
    \label{tab: apllication_theory_chat_result}
\end{table*}

\subsubsection{MathBench-A}


Among all models evaluated in the MathBench application, GPT-4o (GPT-4o-2024-05-13) achieves the highest overall average score, particularly excelling in the more challenging \textit{Middle}, \textit{High}, and \textit{College} stages. Following GPT-4o, Claude-3-Opus and DeepSeek-V2-API outperform in basic arithmetic operations, specifically in the \textit{Arithmetic} and \textit{Primary} stages respectively.
For open-source LLMs, Qwen1.5-110B-Chat stands out as the best performer, distinguishing itself as the leading player among all open-source models. Additionally, DeepSeek-Math-7B-RL, an LLM designed for mathematical tasks, secures its position as the top open-source model in mathematics, despite its relatively small parameter size.

Among open-source chat models, 
performances across models with $\sim$7B, $\sim$20B, and $\sim$70B parameter size reveal distinct capabilities:

\noindent \textbf{$\sim$7B Chat Models. }
InternLM2-Chat-7B and Llama-3-8B-Instruct emerges as the superior model at the $\sim$7B scale and outperforms other 7B Chat models across all stages.
It's noteworthy that, as the difficulty of problems increases, the gap between Llama-3-8B-Instruct and other models also grows. 
For instance, on the five stages from \textit{Arithmetic} to \textit{College} Math, 
It outperforms ChatGLM3-6B by 43.95\%, 73.17\%, 82.48\%, 258.49\%, and 723.53\%, respectively.
The trend indicates that as the difficulty escalates, the performance disparity between models significantly increases since higher-stage math problems often involve more complex concepts and problem-solving strategies, imposing greater demands on the models' comprehension and reasoning abilities. 
All $\sim$7B models struggle with advanced mathematical problems, 
indicating a challenge in smoothly resolving complex questions for small-scale LLMs.

\noindent \textbf{$\sim$20B Chat Models. }
InternLM2-Chat-20B performs the best at the $\sim$20B scale, followed by Qwen-14B-Chat. 
Though Yi-34B-Chat has a much larger parameter size, 
it lags behind other $\sim$20B models. 
Similar to $\sim$7B models, models around $\sim$20B also struggle with more complex mathematical problems at the \textit{High School} and \textit{College} stage.

\noindent \textbf{$\sim$70B Chat Models and Math Models. }
In the realm of large-scale open-source language models, a significant performance disparity is evident when comparing models of varying sizes. Notably, the Qwen1.5-110B-Chat model demonstrates exceptional proficiency in addressing mathematical application problems. Its performance not only surpasses that of other open-source chat-oriented models but also eclipses the capabilities of numerous specialized mathematical models. Remarkably, it exhibits comparable effectiveness to closed-source models, such as GPT-4-0125-Preview, in solving application problems (58.4 vs 58.8).

Focusing on models dedicated to mathematical tasks, the DeepSeek-Math-7B-RL model stands out for its adeptness in tackling application-based questions across a spectrum of stages, encompassing basic \textit{Primary}, \textit{High} and \textit{College} math. Remarkably, it outstrips not only its counterparts, but also the substantially larger DeepSeek-67B-Chat model, by a margin of 24.8\%. This is particularly noteworthy given that the DeepSeek-Math-7B-RL achieves this superior performance with a model size nearly one-tenth that of the DeepSeek-67B-Chat, underscoring the efficiency and targeted capability of the former in mathematical problem-solving domains.


\subsubsection{MathBench-T}


In the theoretical segment of MathBench, designated as MathBench-T, GPT-4o consistently achieved balanced and exceptional results across nearly all theoretical stages. Although Qwen-1.5-110B-Chat exhibited slightly superior performance in the Primary stage, GPT-4o attained an average theoretical score of 87.0. This score was the highest among all tested Closed-source Models and Open-source Chat Models. When combined with an application score of 70.9 in MathBench-A, these results indicate that GPT-4o demonstrates both balanced and superior performance in theory and application on MathBench. This underscores GPT-4o’s strong grasp of theoretical knowledge and its proficiency in applying such knowledge effectively.

Among other models except GPT-4o, the Qwen series models stood out, with Qwen-Max-0428 and Qwen1.5-110B-Chat ranking just behind GPT-4o. Notably, in the theoretical stage of \textit{Primary}, Qwen1.5-110B-Chat scored the highest among all models with an 93.4 CE score. However, GPT-4o's advantage lies in higher educational stages or perhaps more advanced theoretical stages. For example, in the college-level theoretical knowledge stage, GPT-4o achieved a CE score of 84.1, which is 16.9 points higher than the best open-source math model, Deepseek-Math-7B-RL.

Similar to MathBench-A, InternLM2-Chat-7B demonstrated robust theoretical capabilities at the common 7B stage models. Despite achieving similar effectiveness to Llama-3-8B-Instruct, InternLM2-Chat-7B exhibited a significantly larger lead in the theoretical stage, surpassing the Qwen-7B-Chat model by 31.3\%. Within the domain of mathematical chat models, Deepseek-Math-7B-RL continued to outperform numerous mathematical models, achieving superior results in both theory and application. Notably, it even surpassed Llama-3-70B-Instruct in the theoretical domain.

Overall, in the tests conducted on MathBench, there was not a significant rank change between models in terms of theoretical and application capabilities. That is, models that ranked highly in application capabilities also tended to perform well in theoretical tests, and vice versa.


\begin{figure}[!t]
    \centering
    \includegraphics[width=1\linewidth]{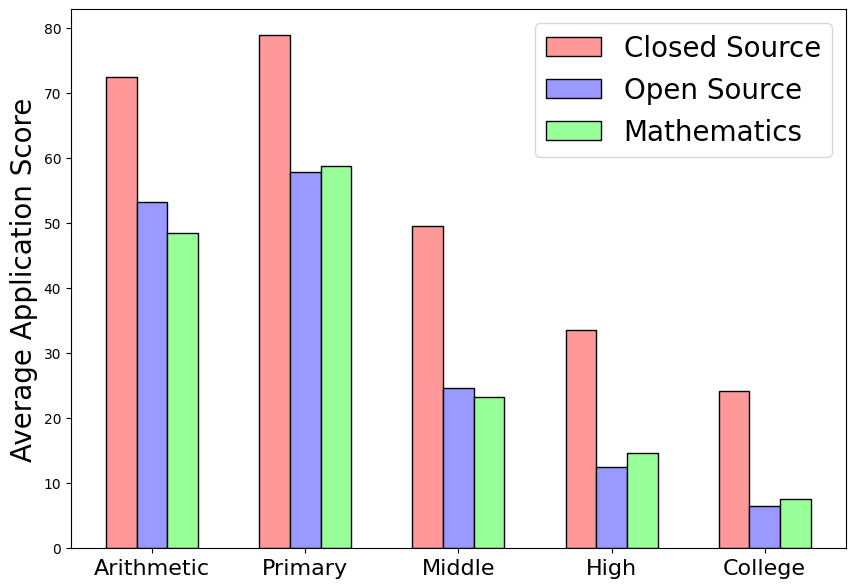}
    \caption{\textbf{Scores of Application Problems at Each Stage}. 
    Models exhibit similar performances in \emph{Arithmetic} and \emph{Primary} stages, while demonstrating a clear performance decline from \emph{Primary} to \emph{College} stages.}
    \label{fig:application_average}
\end{figure}

\subsection{Evaluation of Base Models}
\begin{table*}[!htb]
    \begin{minipage}{.55\linewidth}
      \centering
        \resizebox{\linewidth}{!}{

\begin{tabular}{lccccc|c}
\hline
\textbf{Models} & \textbf{Arith} & \textbf{Primary} & \textbf{Middle} & \textbf{High} & \textbf{College} & \textbf{Avg.} \\
\hline
\multicolumn{7}{c}{\textit{\bluecolor{$\heartsuit$Open-source Base Models}}} \\
Llama-2-7B & 28.0 & 9.0 & 27.0 & 31.3 & 31.7 & 25.4 \\
Deepseek-7B-Base & 31.0 & 14.0 & 26.7 & 32.3 & 28.0 & 26.4 \\
Baichuan2-7B-Base & 44.0 & 24.3 & 31.0 & 33.7 & 28.7 & 32.3 \\
Mistral-7B-v0.1 & 42.7 & 30.0 & 35.0 & 32.7 & 35.3 & 35.1 \\
Qwen-7B & 44.3 & 46.3 & 38.0 & 32.3 & 36.3 & 39.5 \\
ChatGLM3-6B-Base & 39.7 & 48.3 & 43.7 & 38.0 & 33.0 & 40.5 \\
InternLM2-7B & \underline{49.0} & \underline{63.3} & \underline{46.7} & \underline{38.7} & \underline{38.0} & \underline{47.1} \\
\hline
Llama-2-13B & 30.0 & 21.0 & 30.7 & 31.7 & 28.3 & 28.3 \\
Baichuan2-13B-Base & 47.7 & 42.3 & 36.7 & 31.7 & 38.7 & 39.4 \\
Qwen-14B & 52.0 & 57.7 & 51.7 & 39.3 & 43.7 & 48.9 \\
InternLM2-20B & \underline{57.3} & \underline{70.7} & \underline{45.3} & \underline{39.3} & \underline{37.0} & \underline{49.9} \\
\hline
Llama-2-70B & 44.3 & 50.3 & 35.3 & 34.0 & 40.7 & 40.9 \\
Mixtral-8x7B-v0.1 & 55.3 & 49.7 & 35.0 & 34.0 & 42.3 & 43.3 \\
Deepseek-67B-Base & 45.3 & 62.7 & 41.3 & 40.3 & 41.7 & 46.3 \\
Qwen-72B & \textbf{62.3} & \textbf{71.7} & \textbf{62.0} & \textbf{58.0} & \textbf{51.3} & \textbf{61.1} \\
\multicolumn{7}{c}{\textit{\greencolor{$\triangle$Mathematical Models}}} \\
Llemma-7B & 41.3 & 27.3 & 34.7 & 41.3 & 41.0 & 37.1 \\
Llemma-34B & 44.3 & 45.0 & 35.7 & 34.0 & 40.3 & 39.9 \\
InternLM2-Base-Math-7B & 46.0 & 42.0 & 43.0 & 35.3 & 38.7 & 41.0 \\
InternLM2-Base-Math-20B & 48.0 & 50.3 & 46.3 & 42.0 & 40.3 & 45.4 \\
Deepseek-Math-7B-Base & \underline{58.3} & \underline{62.0} & \underline{47.0} & \underline{47.0} & \underline{47.7} & \underline{52.4} \\
\hline
\end{tabular}
        }
        \caption*{\small{MathBench-A.}}
    \end{minipage}%
    \hfill
    \begin{minipage}{.5\linewidth}
      \centering
        
        \resizebox{\linewidth}{!}{
\begin{tabular}{lcccc|c}
\hline
\textbf{Models} & \textbf{Primary} & \textbf{Middle} & \textbf{High} & \textbf{College} & \textbf{Avg.} \\
\hline
\multicolumn{6}{c}{\textit{\bluecolor{$\heartsuit$Open-source Base Models}}} \\
Llama-2-7B & 13.6 & 5.1 & 6.8 & 8.4 & 8.5 \\
Deepseek-7B-Base & 24.2 & 17.7 & 12.6 & 17.9 & 18.1 \\

Baichuan2-7B-Base & 35.2 & 30.5 & 20.1 & 25.6 & 27.9 \\
Qwen-7B & 38.0 & 36.8 & 24.3 & 26.1 & 31.3 \\
Mistral-7B-v0.1 & 39.8 & 33.4 & 27.8 & 45.9 & 36.7 \\
ChatGLM3-6B-Base & \underline{53.0} & \underline{48.6} & 37.1 & 37.7 & 44.1 \\
InternLM2-7B & 49.0 & 45.9 & \underline{43.6} & \underline{50.5} & \underline{47.2} \\
\hline
Llama-2-13B & 29.9 & 21.4 & 17.8 & 22.8 & 23.0 \\
Baichuan2-13B-Base & 47.2 & 41.9 & 28.5 & 39.9 & 39.4 \\
InternLM2-20B & 52.9 & 48.7 & 46.1 & 54.9 & 50.6 \\
Qwen-14B & \underline{68.4} & \underline{63.0} & \underline{52.3} & \underline{55.7} & \underline{59.9} \\
\hline
Llama-2-70B & 48.0 & 42.9 & 35.6 & 53.0 & 44.9 \\
Mixtral-8x7B-v0.1 & 55.5 & 49.1 & 46.1 & 61.4 & 53.0 \\
Deepseek-67B-Base & 65.7 & 61.9 & 47.7 & 59.8 & 58.8 \\
Qwen-72B & \textbf{84.1} & \textbf{77.5} & \textbf{70.9} & \textbf{77.5} & \textbf{77.5} \\
\multicolumn{6}{c}{\textit{\greencolor{$\triangle$Mathematical Models}}} \\
Llemma-7B & 24.3 & 26.6 & 22.7 & 35.8 & 27.3 \\
InternLM2-Chat-Math-7B & 34.4 & 37.7 & 34.1 & 48.4 & 38.7 \\
Llemma-34B & 42.7 & 45.2 & 40.9 & 52.8 & 45.4 \\
InternLM2-Base-Math-20B & 48.7 & 48.4 & 46.0 & 53.2 & 49.1 \\
Deepseek-Math-7B-Base & \underline{62.5} & \underline{64.5} & \underline{54.2} & \underline{67.4} & \underline{62.1} \\
\hline
\end{tabular}
        }
        \caption*{\small{MathBench-T.}}
    \end{minipage}
    \caption{\textbf{Overall Comparison of Base Models on MathBench A \& T}. The \textit{Arithmetic} and \textit{Primary} stage for MathBench-T are combined because they share the same theory knowledge. Models are classified into categories based on their parameter size and the dataset they were trained on. The model name in \textbf{bold} indicates the top performer within all base models, while an \underline{underline} signifies the leading model within a similar parameter size group.}
    \label{tab: application_theory_base_result}
\end{table*}
The results for the Base models are presented in \Cref{tab: application_theory_base_result}. Consistency in performance is observed between the Base models and their Chat model counterparts, with InternLM2-7B emerging as the optimal model in the 7B parameter range. Qwen-14B and Qwen-72B demonstrate superior performance within their respective parameter classes on the MathBench benchmark. For mathematical tasks, Deepseek-Math-7B-Base's results align closely with those seen in the Chat model evaluations, indicating a significant correlation between the efficacy of Base models and their corresponding Chat models, which leads to similar performance trends across models within the same category.

Notably, ChatGLM3-6B-Base secures the second-highest ranking in the 7B base model evaluation, outperforming several other models, including Qwen-7B and Mistral-7B-v0.1. However, this performance is not mirrored in its Chat model variant, ChatGLM3-6B, which is surpassed by Qwen-7B-Chat by 95.2\% on MathBench-A and by 104.7\% on MathBench-T. This discrepancy in performance between the Chat and Base versions of the model may be attributed to the different fine-tuning strategies applied during the subsequent tuning phase, which could explain the observed gap in performance.

\subsection{Detailed Analysis}
\begin{figure*}
    \centering
    \includegraphics[width=1\linewidth]{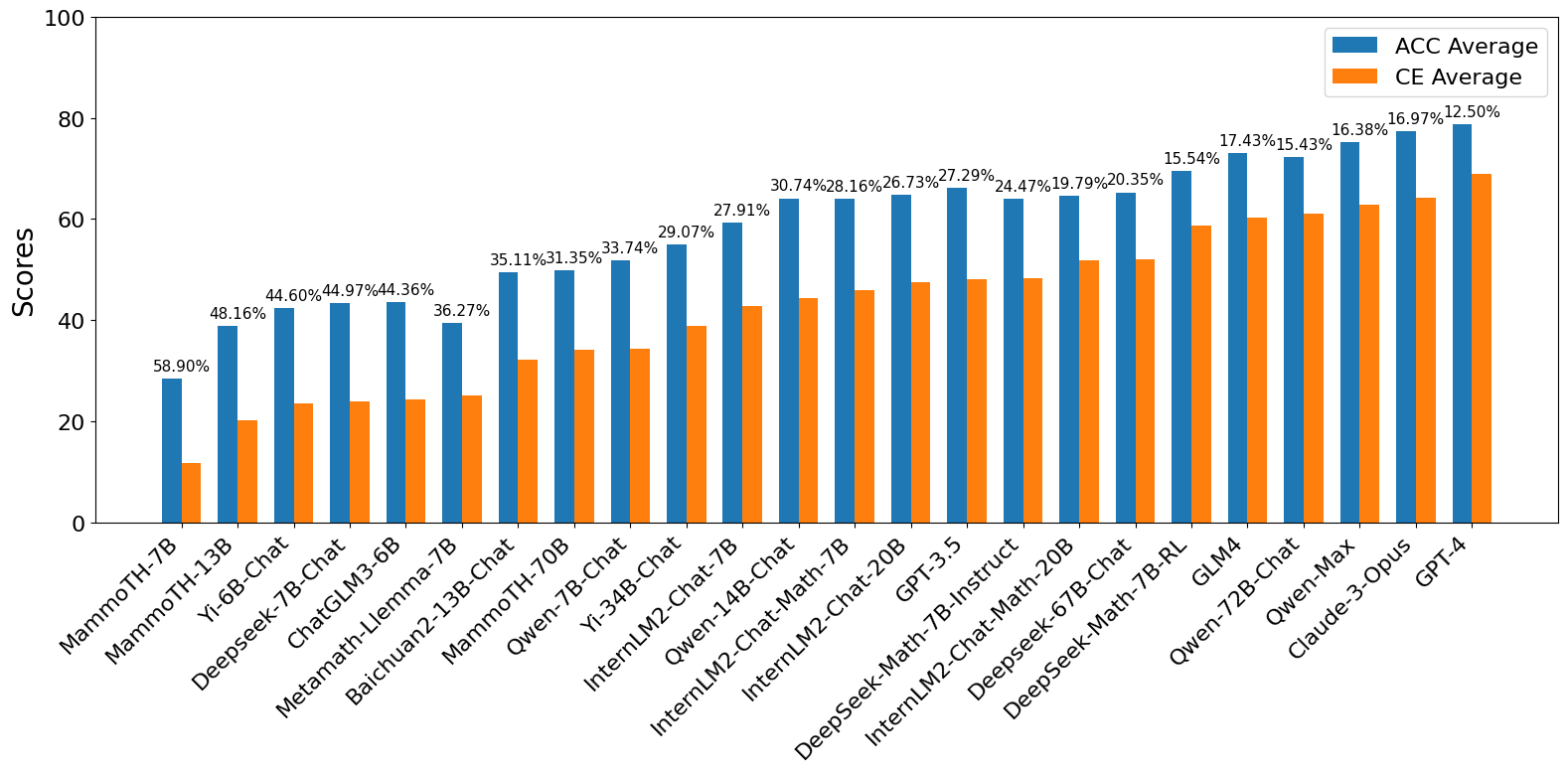}
    \caption{\textbf{CE Evaluation vs. ACC Evaluation}. The ACC evaluation queries the model once per question and checks for correctness, whereas the CE (CircularEval) conducts a more stringent and robust assessment by rolling out evaluations four times with shuffled answer options, deeming a question correct only if all attempts are accurate. The percentages depicted in the figure represent the performance decrease of models in the CE evaluation compared to the ACC evaluation.}
    \label{fig:acc ce comparison}
\end{figure*}

With MathBench, 
we can easily assess the model's mathematical capabilities at different granularities including education stage, language, subject area, or even specific topics with questions on both theories and applications.
Below, we will delve deeper into the evaluation results and discuss about the following questions: 

\paragraph{The Gap between Circular and Accuracy Evaluation}
A comparison between Circular Evaluation (CE) scores and Accuracy (ACC) scores is illustrated in \Cref{fig:acc ce comparison}. As model performance improves, the discrepancy between CE and ACC scores becomes increasingly narrow, suggesting that more powerful models tend to provide more robust and stable answers in mathematical question answering.

\begin{figure}[!ht]
\centering
\includegraphics[width=1\linewidth]{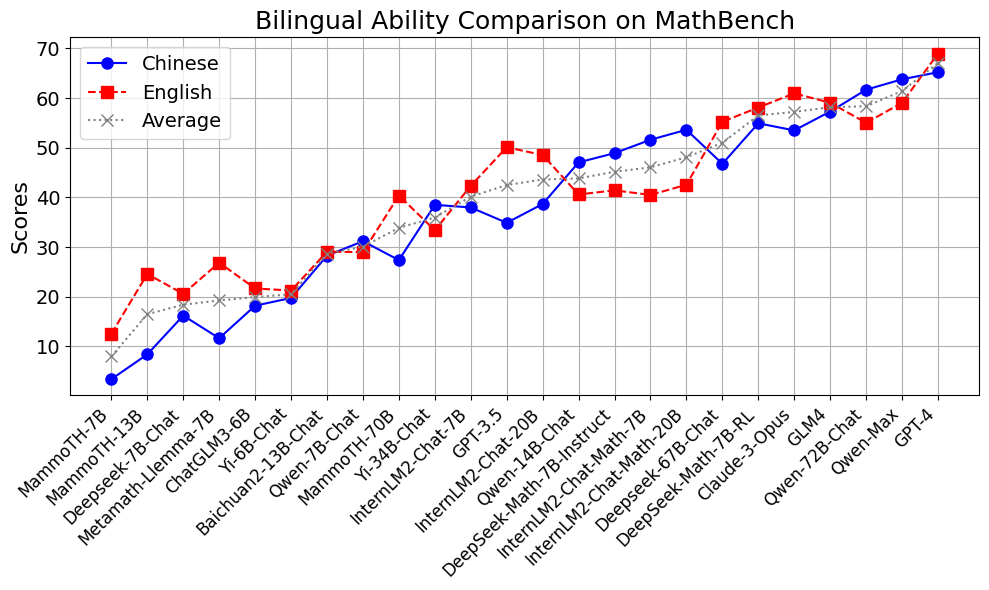}
\caption{\textbf{Bilingual Comparison on MathBench}. showcasing scores in Chinese, English , and their average for the gray dashed line. The \textit{Arithmetic} stage is not include because there no impact of language in it.}
\label{fig:bilingual-ability}
\end{figure}

\paragraph{How Models' Scores on Application Problems Vary Across Stages?}
Figure \ref{fig:application_average} presents the average performance of all aforementioned models on application questions in MathBench. 
Most models perform reasonably well on \textit{Arithmetic} and \textit{Primary} math problems. 
However, their effectiveness drastically declines when it comes to the \textit{Middle} stage or above. 
Such phenomenon suggests that existing models are good at tasks that can be solved through direct computation, pattern recognition, or memorizing basic concepts.
However, they showcase inferior performance when solving more complex math problems.

\paragraph{Is There A Gap between Theory Understanding and Application Capabilities?} 

\begin{figure*}
    \centering
    \includegraphics[width=1\linewidth]{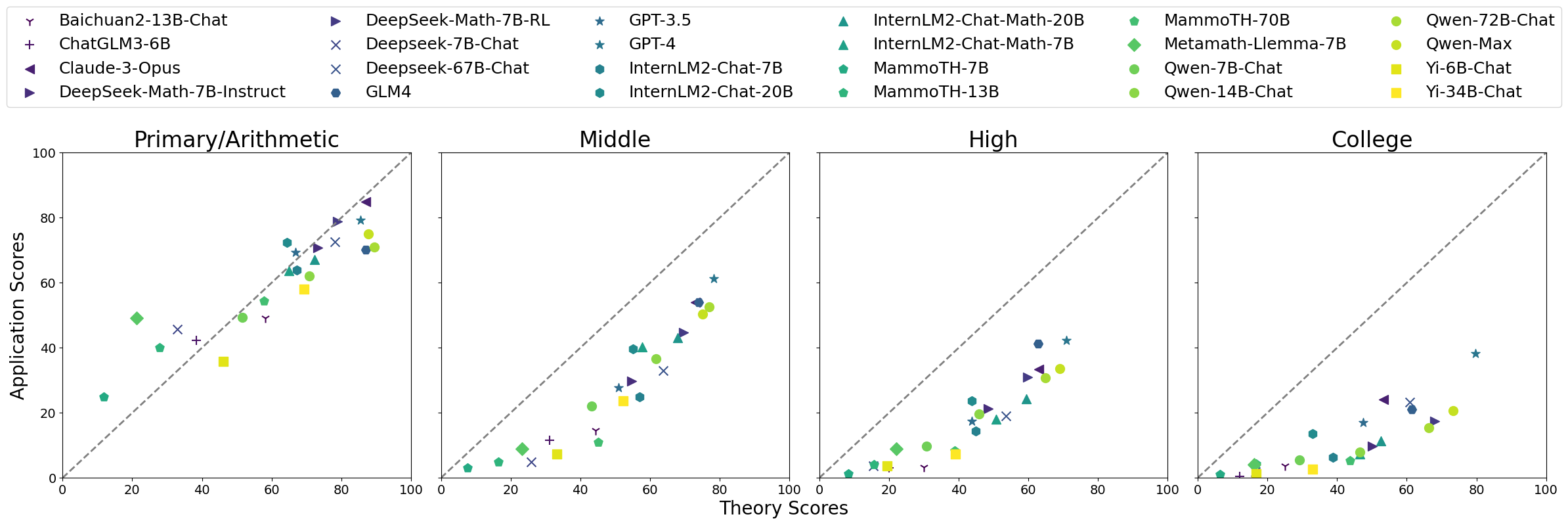}
    \caption{\textbf{Theoretical Score vs. Applied Score on MathBench}. \textit{Primary} and \textit{Arithmetic} are averaged because they share the same theory knowledge points.}
    \label{fig:app_theory_with_stages}
\end{figure*}

Theories serve as the foundation for addressing the majority of application problems. 
As illustrated in \Cref{fig:app_theory_with_stages}, 
we present the trend of LLMs in terms of theoretical and application scores across different stages. 
In the \textit{Primary} stage, the two scores are highly correlated for most LLMs, with only a few exceptions. 
Among top-ranked models, Qwen-72B-Chat demonstrates the best theoretical ability, while Claude-3-Opus demonstrates superior application ability. 
When it comes to more advanced stages, 
models require better computational and reasoning capabilities to achieve good application scores. 
GPT-4 leads in the application track across all stages,
while the gap is larger in more advanced stages. 
For example, 
comparing to Qwen-72B-Chat,
the difference in theoretical and application scores $(D_t,D_a)$ increases from $(1.4, 8.7)$ in the \textit{Middle} stage to $(6.0, 11.7)$ in the \textit{High} stage, and finally to $(13.5, 23.0)$ in the \textit{College} stage. 
Moreover, from the \textit{Middle} stage onwards, there is a general trend of decline in both theoretical and application abilities of models. 
Compared to theoretical scores, the decline in application scores is more serious. 


\begin{figure*}[!th]
    \centering
    \includegraphics[width=1\linewidth]{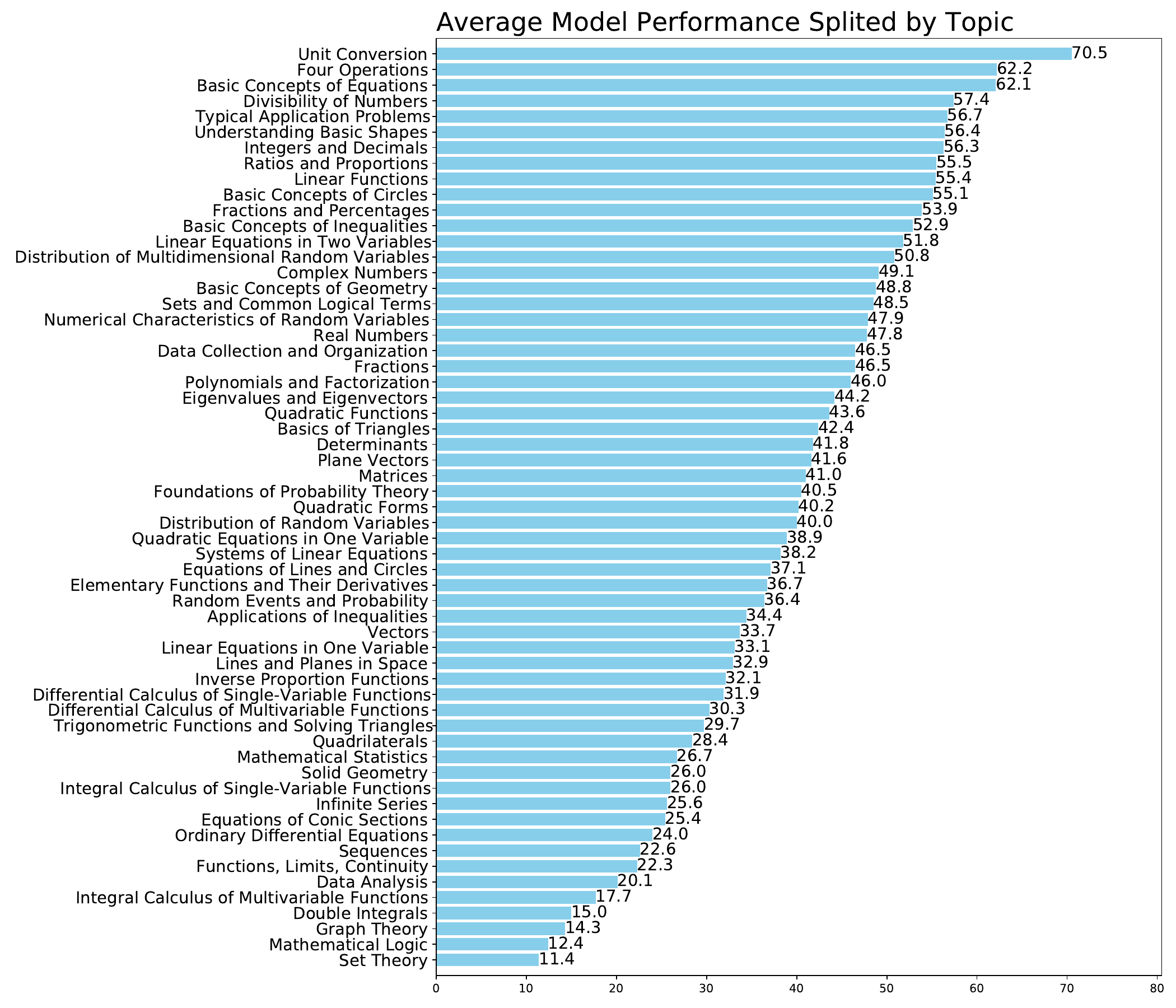}
    \caption{\textbf{Average Model Scores with Topics}. We average the scores of every Chat model for each topic in MathBench. The models excel at basic-level problems, such as single Unit Conversion and basic Four Operations, but as the required reasoning and computational abilities for a topic increase, the performance of the models gradually declines, as observed in topics like Double Integrals, Set Theory, and Mathematical Logic.}
    \label{fig:average topic}
\end{figure*}

\paragraph{Detailed Model Performance for Each Topic}
In \Cref{fig:average topic} indicating average knowledge point performance, it is evident that topics associated with fundamental mathematical skills—such as 'Unit Conversion,' 'Four Operations,' and 'Basic Concepts of Equations'—register higher average scores. This suggests that the majority of models exhibit a proficient command of simple and elementary mathematical questions.

Conversely, topics demanding abstract reasoning and intricate computations, like 'Double Integrals,' 'Mathematical Logic,' and 'Set Theory,' show lower average scores. Addressing the mathematical queries in these topics may require bespoke model analysis and optimization. It is crucial to pinpoint the source of inaccuracies within these topics, whether it be due to a deficit in reasoning ability or an unstable grasp of the relevant foundational theoretical concepts.

\paragraph{Which Model Performs Better under the Bilingual Scenario?}
\Cref{fig:bilingual-ability} demonstrates the bilingual capabilities of various LLMs on MathBench, indicating the importance of linguistic versatility in mathematical tasks that demand an understanding of nuances in language and math concepts across different languages.
Among all LLMs, GPT-4 leads with the highest bilingual score of 67.1, showing a balanced performance between Chinese (65.2) and English (69.0). This demonstrates GPT-4's advanced bilingual processing abilities.
Other models including Qwen-72B-Chat and DeepSeek-Math-7B-RL also exhibit significant bilingual capabilities. 
It's also noteworthy that among all LLMs evaluated, most of them feature a much larger performance gap between Chinese and English, compared to GPT-4. The detailed result of bilingual test of MathBench can be found in \Cref{app: bilingual results}.

\paragraph{Enhancing Model Proficiency in Fundamental Theories: Initial Explorations}
In an effort to augment the model's grasp and application of theoretical concepts in problem-solving contexts, we embarked on exploratory initiatives, focusing primarily on two methodologies: Chain of Thought (CoT) and Knowledge Infusion.

We selectively sampled 200 questions from MathBench, deliberately skewed towards theoretical reasoning and application (with a distribution of 40\% application-oriented and 60\% theory-centric questions), to perform ancillary experiments on the Qwen-72B-Chat model. The outcomes, delineated in Table~\ref{tab:enhance concept}, elucidate the accuracy achieved through different strategic approaches.

\begin{itemize}
    \item \textbf{Straight:} Immediate response without preceding CoT.
    \item \textbf{CoT:} Response derived post-CoT, serving as MathBench's standard evaluative criterion.
    \item \textbf{Straight-Knowledge:} Immediate response, preconditioned by the integration of relevant knowledge points prior to posing the question.
    \item \textbf{CoT-Knowledge:} Response post-CoT, facilitated by the preliminary inclusion of pertinent knowledge points.
\end{itemize}

\begin{table}[h!]
\centering
\begin{tabular}{|l|c|}
\hline
\textbf{Strategy}           & \textbf{Accuracy (\%)} \\ \hline
Straight           & 26.6     \\ 
CoT                & 29.8     \\ 
Straight-Knowledge & 31.3     \\ 
CoT-Knowledge      & 33.4     \\ \hline
\end{tabular}
\caption{Comparative accuracy of different strategies.}
\label{tab:enhance concept}
\end{table}

Knowledge points were meticulously curated from academic textbooks and instructional resources. The experimental data suggests a progressive enhancement in efficacy: Straight < CoT < Straight-Knowledge < CoT-Knowledge. This progression evidences the significant impact of both CoT and knowledge point infusion on augmenting model performance for questions heavily reliant on theoretical reasoning or practical application, with their combined utilization yielding the most favorable outcomes.

\section{Discussion}

\begin{figure}
    \centering
    \includegraphics[width=1\linewidth]{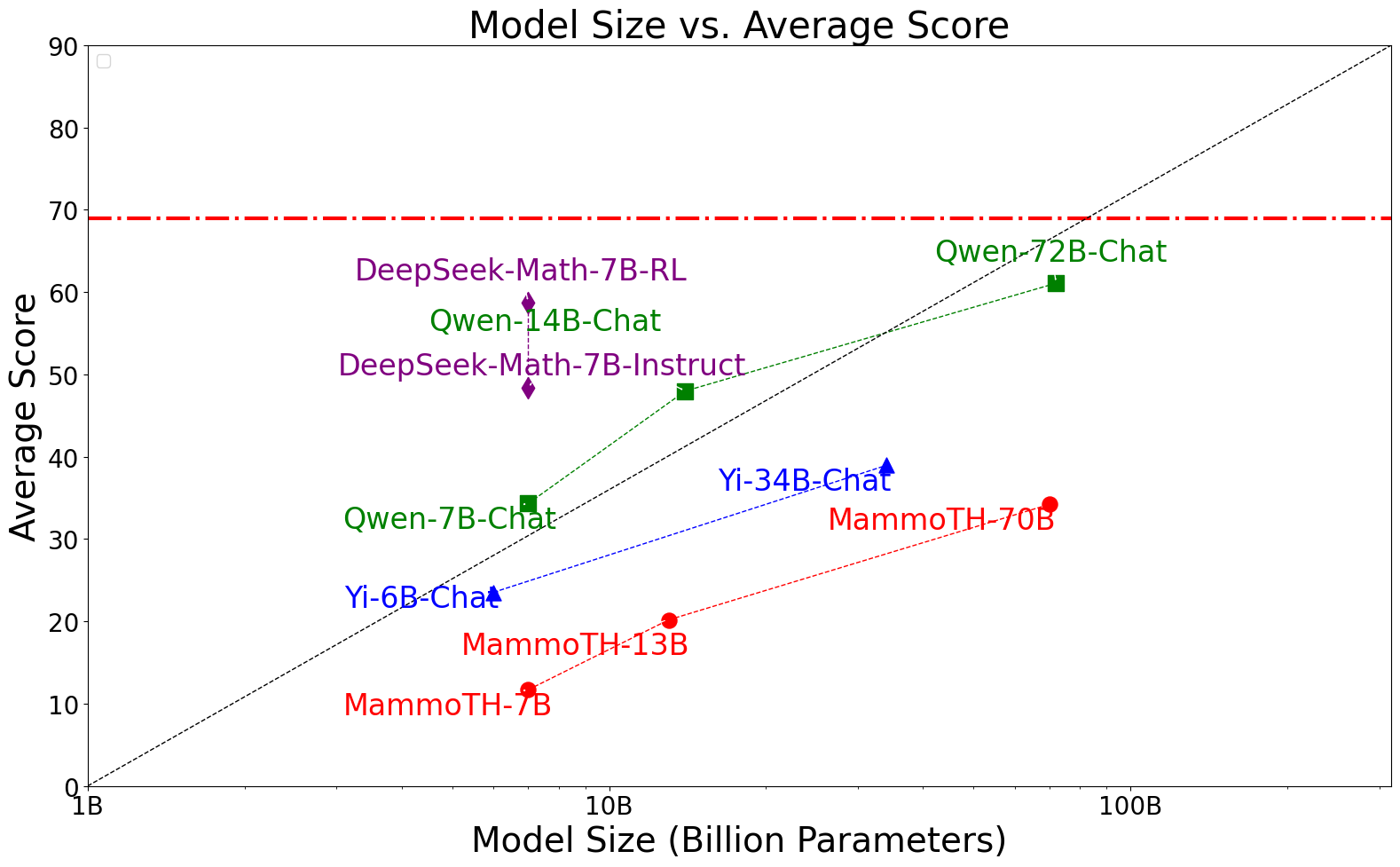}
     \caption{\textbf{Model Size vs. Average Score}. The comparison chart of model parameter size versus performance on MathBench for selected representative models, with models from the same series connected by lines of the same color. The horizontal red dotted line represents the score of GPT-4.} 
    \label{fig:scale_score}
\end{figure}


\subsection{How Models Perform with Code Agent on MathBench}
\begin{figure*}[ht]
    \centering
    \begin{subfigure}[b]{0.49\linewidth}
        \includegraphics[width=\linewidth]{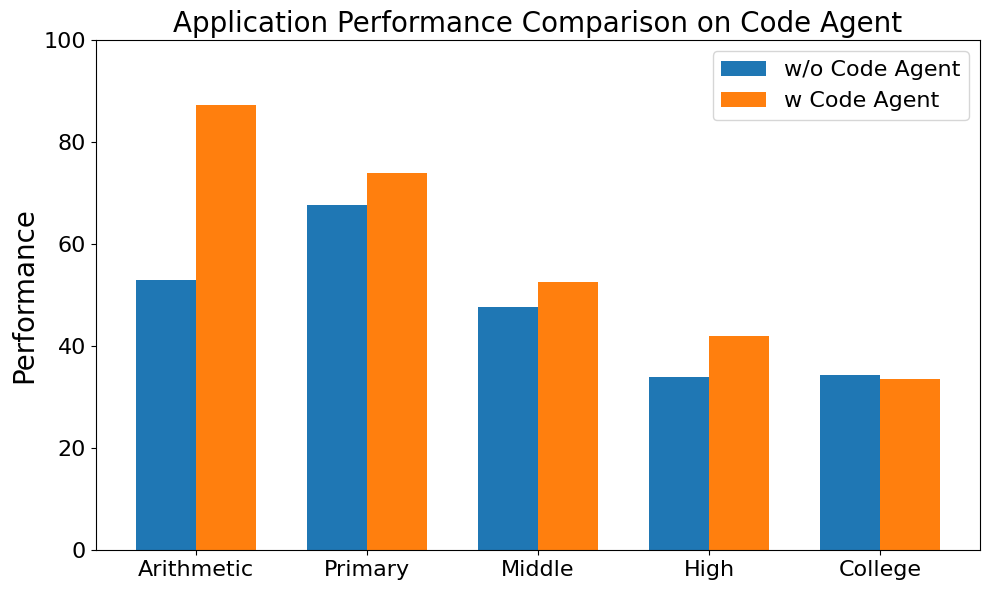}
        \label{fig:code_agent1}
    \end{subfigure}
    \hfill 
    \begin{subfigure}[b]{0.49\linewidth}
        \includegraphics[width=\linewidth]{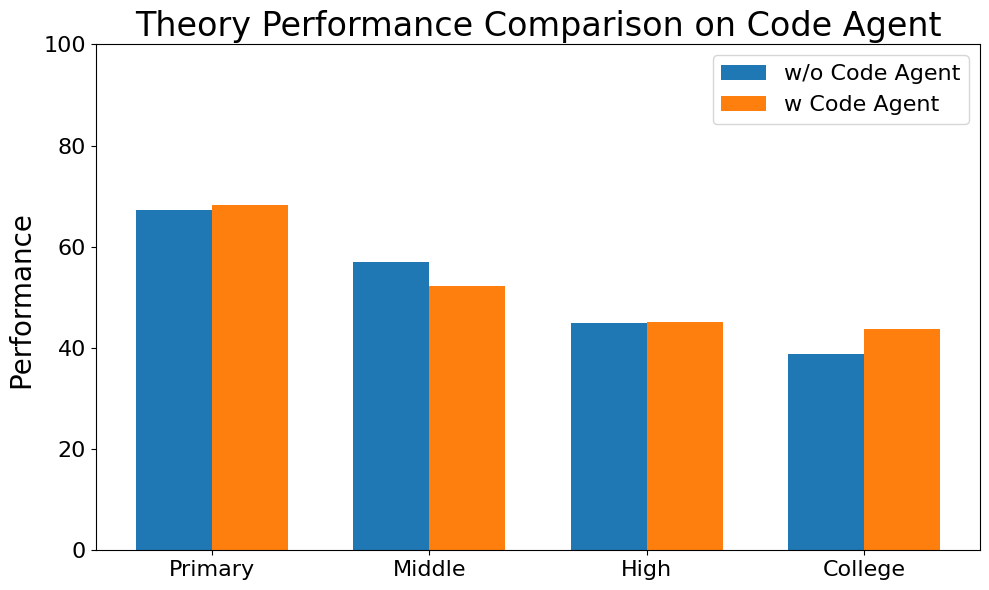}
        \label{fig:code_agent2}
    \end{subfigure}
    \caption{\textbf{Code Agent Performance on MathBench}. We use InternLM2-7B-Chat for the comparison and the results are divided into two parts with stages: Application and Theory. }
    \label{fig: code_agent_results}
\end{figure*}

We utilize the external code interpreter and follow the ReAct~\citep{yao2023react} protocol in Lagent~\citep{lagent2023} to evaluate LLMs' ability in solving mathematical problems of MathBench. The results, as depicted in \Cref{fig: code_agent_results}, show the comparison of performance with and without the Code Agent on the Theory and Application sections of MathBench. Overall, the inclusion of the Code Agent significantly enhances performance in the Application section, especially in Arithmetic, where it boosts the performance of InternLM2-7B-Chat by about 64\% (from 53.0 to 87.3). This demonstrates that the addition of the Code Agent can substantially improve the model's basic numerical calculation capabilities. However, for more complex problems, such as those in the College level Application section, the Code Agent does not notably improve model capabilities and even slightly degrades performance. For theoretical problems, the Code Agent does not significantly enhance InternLM2-7B-Chat's performance across various stages on MathBench. This suggests that mathematical theoretical ability, as a crucial foundational skill for models, requires more than just external tools. Instead, it necessitates exploring more effective ways to enhance large language models' understanding and application of mathematical concepts.

\subsection{Effect of Model Size on Math Capabilities}
We found that for models of different sizes within the same series, 
most of them conform to the Scaling Law~\citep{kaplan2020scaling} on MathBench.
For example, Qwen series, MammoTH series, and Yi series have shown steady improvement in their MathBench scores as the parameter size increases, 
as shown in \Cref{fig:scale_score}. 
However, it doesn't mean that models with small parameter sizes can not achieve good math performance.
For instance, DeepSeek-Math-7B demonstrates outstanding performance on MathBench and outperforms models with 10x parameters, including DeepSeek-72B and a larger math model MammoTH-70B. 

\subsection{Error Analysis}
\begin{figure}
    \centering
    \includegraphics[width=1\linewidth]{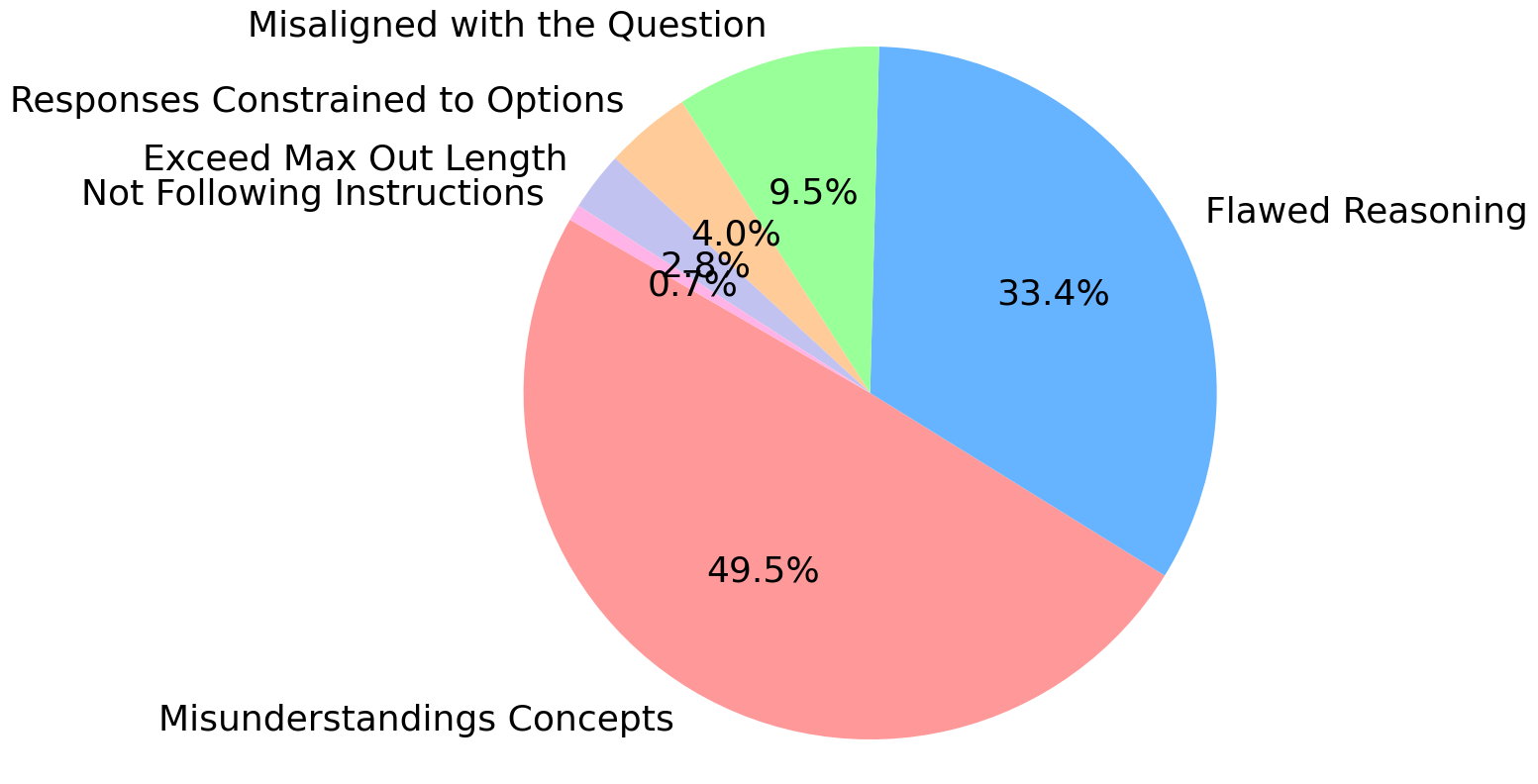}
    \caption{\textbf{Response Error Analysis for Both Theoretical and Application Questions. } The predominant sources of errors are a fundamental misunderstanding of the concepts, followed by incorrect reasoning paths.}
    \label{fig:response-error-analysis}
\end{figure}

In our study, we conduct a comprehensive error analysis on a set of 80 theoretical and 100 application questions random selected from every stages, for models selected across different scales, as illustrated in \Cref{fig:response-error-analysis}. 
The error categories are uniformly observed across all evaluated models, indicating common challenges that transcend specific parameter scales.
Our selection of models includes GPT-3.5, GPT-4, InternLM2-Chat-7B, Qwen-14B-Chat, Qwen-72B-Chat, Deepseek-Math-7B-RL and MammoTH-70B. Detailed cases for error analysis can be found in \Cref{app: error analysis}.

\noindent \textbf{Insufficiency of knowledge. } 
For theoretical questions, 78\% of model errors are due to misconceptions about mathematical concepts, 
which notably emerged as a significant concern in several models. 
Such errors accounted for 49.5\% of all mistakes, 
underscoring a general challenge in grasping fundamental knowledge and terminology.

\noindent \textbf{Deficiencies in reasoning. } 
Furthermore, models exhibited shortcomings in logical reasoning, with 33.4\% of errors attributed to logically consistent but flawed reasoning processes.
Moreover, errors such as reasoning that deviated from the intended query accounting for 9.6\%, underscored the models' limitations in understanding user intentions and providing pertinent responses. 
We also notice that errors related to reasoning increased with task difficulty.

\noindent \textbf{Response length limit. } 
Though statistically not the primary error mode (4.0\%), 
responses that exceeded the token limit shed light on the challenge of reasoning complex tasks within limited length and adhering to given instructions.

\noindent \textbf{Other cases. } 
Occasionally, models will generate responses devoid of an explicit reasoning process, 
obstructing additional scrutiny. 
Moreover, models endowed with enhanced reasoning capabilities exhibit a greater capacity for critical thinking regarding the options presented, 
thereby offering alternative answers that transcend the limitations of predetermined choices.


\subsection{Reasoning Path}
Analyzing the reasoning paths of various models across multiple difficulty levels reveals significant performance disparities. 
We set a brief discussion below and provide more detailed cases for reasoning path analysis in \Cref{app: reasoning path}.

\noindent \textbf{Performance across diverse difficulties. } 
In straightforward scenarios, models swiftly solve the problems with direct reasoning and yield logical outcomes. 
Yet, complex issues, marked by dense symbols, vast knowledge, and intricate links, necessitate broader knowledge navigation, accentuating divergences in deductive strategies.

\noindent \textbf{Reasoning paths of chat models with different parameter sizes. } 
Small-scale chat models strive for logical coherence in mathematics, 
yet may make mistakes due to knowledge deficiencies, 
particularly in symbol interpretation and relational understanding. 
In contrast, large-scale models feature expansive knowledge and nuanced insights, which enhance symbol processing and minimizing knowledge gaps. 
However, even with substantial parameters, challenges in efficient knowledge  management persist, 
occasionally leading to irrelevant diversions and diminished reasoning efficacy.

\noindent \textbf{Reasoning paths of math models. } 
Specialized math models, despite the smaller parameter sizes, exhibit superior mathematical comprehension and systematic logical reasoning. 
They excel in applying mathematical knowledge and notation to reason through complex problems. 

\noindent \textbf{Superlative deductive navigation of Closed-source models. } 
GPT-4 stands out for its effective reasoning and deep problem comprehension. 
It engages in logical, coherent, and succinct discussions, adeptly navigate complex reasoning paths, and manage mathematical symbols effectively. GPT-4 distinctively recognizes problem statement ambiguities, showcasing a detailed and nuanced reasoning process.

\section{Related Work}
Solving math word problems through automated methods has been a long-standing concern for researchers. This section summarizes seminal studies and delineates key evaluation datasets proposed for assessing mathematical problem-solving approaches, tracing the field's evolution from its origins to the present day.
\paragraph{Preliminary Mathematical Datasets}
Previous works proposed datasets such as Alg514~\citep{Alg514}, SingleEq~\citep{SingleEq}, and DRAW-1K~\citep{draw} are primarily concentrated on elementary linear algebraic problems. Similarly, datasets like AddSub~\citep{AddSub} and SingleOp~\citep{SingleOp} MultiArith~\citep{MultiArith} are exclusively dedicated to fundamental arithmetic operations: addition, subtraction, multiplication, and division. These datasets are very limited both in the form and content of their assessments, focusing solely on a specific small part of basic mathematics.

\paragraph{Benchmarks tailored to specific educational tiers} 
Some benchmarks are designed based on educational levels. Math23k~\citep{Math23k} collects a corpus of real math word problems for elementary school students. While ASDiv~\citep{ASDiv} expands the textual patterns to encompass most problem types found in elementary mathematics. GSM8K~\citep{gsm8k} presents a high-quality collection of elementary mathematical word problems that, on average, require multiple steps to solve and provide solutions in natural language annotations. These datasets mostly focus on elementary mathematics and seldom examine college-level knowledge.
\paragraph{Enriching the diversity of mathematical problem types within benchmarks}
MathQA~\citep{MathQA} seeks to categorize problems from AQuA~\citep{AQuA} into different mathematical domains based on the frequency of mathematical terminology used. Mathematics Dataset~\citep{Mathematics_Dataset} expands the subject of mathematics and this dataset covers a broader spectrum of mathematics, including arithmetic, algebra, probability, and calculus. MATH~\cite{math} features a higher level of complexity, comprising problems ranging from arithmetic to calculus, and aims at testing models' capabilities in understanding and solving complex mathematical challenges.  
While these efforts have enhanced the diversity of the data in certain aspects, they are lacking in diversity in other aspects such as question formulation~\citep{Mathematics_Dataset}.


\paragraph{Integrating mathematical problems with domain knowledge}


NumGLUE~\citep{NumGLUE}  not only assesses the ability of models to solve mathematical problems given direct computational expressions, but it also designs multiple tasks to comprehensively evaluate the models' abilities to use other reasoning skills, such as common sense and reading comprehension. Lila~\citep{Lila} is developed through the extension of 20 datasets that cover a broad range of mathematical topics. This dataset exhibits varying degrees of linguistic complexity and features diverse question formats as well as background knowledge requirements. These works inspire us to design more diversified testing scenarios.
\section{Conclusion}

In summary, MathBench adopts structured approaches to categorize questions by stage and knowledge level.
It aims to provide a comprehensive evaluation of LLMs' mathematical proficiency. 
By covering a wide range of subject areas and topics across educational stages, MathBench offers a unique resource for researchers and educators interested in advancing the field of mathematical learning and assessment.

\section{Limitations}

We have developed a comprehensive mathematical evaluation benchmark, MathBench, which includes a detailed knowledge framework and multi-dimensional, fine-grained mathematical questions. Despite its strengths, the benchmark currently has several limitations, which are summarized as follows:

\textbf{Data Source:} To enhance diversity, some questions were sourced from open-source datasets($\sim19\%$). However, these open-source questions may be subject to data contamination, which could compromise the assurance that models have not been exposed to these questions before. In future iterations, we plan to automate the construction of questions across various stages to more effectively test the models' genuine mathematical capabilities.


\textbf{Lack of Detailed Reasoning Paths:} Given the diversity of questions and time constraints, MathBench currently does not provide detailed reasoning paths for each question. This limitation makes it challenging to unlock the full potential of the questions. Moving forward, we aim to investigate semi-automated methods to offer both natural language and code-based reasoning approaches for each question, thereby maximizing the value of MathBench's questions.

\section{Ethical Considerations}
For our benchmarks, we relied on reference materials and closed-source models that are accessible to the public, thereby avoiding any potential harm to individuals or groups. The data produced by the LLMs underwent a meticulous human selection and processing phase to ensure the protection of privacy and confidentiality. We did not use any personally identifiable information, and all data were anonymized prior to analysis. Additionally, we employed ChatGPT and Grammarly to refine our manuscript's language.

\section{Acknowledgements}
We thank the OpenCompass team for their diligent efforts in data collection and validation, which ensured the quality and completion of this work. We extend our special thanks to Jiaye Ge for providing comprehensive support on this project, orchestrating the data collection process, and ensuring the timely delivery of the dataset. We also thank OpenDataLab for providing data sources for part of this work. Our gratitude extends to Mo Li for his suggestions on the modifications of some charts in the paper, as well as Kuikun Liu for the Lagent support. Chuyu Zhang and Jingming Zhuo provided valuable advice on the writing of the paper.

This work was supported by National Key R\&D Program of China 2022ZD0161600 and Shanghai Postdoctoral Excellence Program 2022235. 

\bibliography{custom}

\begin{thebibliography}{39}
\expandafter\ifx\csname natexlab\endcsname\relax\def\natexlab#1{#1}\fi

\bibitem[{Achiam et~al.(2023)Achiam, Adler, Agarwal, Ahmad, Akkaya, Aleman, Almeida, Altenschmidt, Altman, Anadkat et~al.}]{achiam2023gpt}
Josh Achiam, Steven Adler, Sandhini Agarwal, Lama Ahmad, Ilge Akkaya, Florencia~Leoni Aleman, Diogo Almeida, Janko Altenschmidt, Sam Altman, Shyamal Anadkat, et~al. 2023.
\newblock Gpt-4 technical report.
\newblock \emph{arXiv preprint arXiv:2303.08774}.

\bibitem[{Amini et~al.(2019)Amini, Gabriel, Lin, Koncel-Kedziorski, Choi, and Hajishirzi}]{MathQA}
Aida Amini, Saadia Gabriel, Peter Lin, Rik Koncel-Kedziorski, Yejin Choi, and Hannaneh Hajishirzi. 2019.
\newblock \href {http://arxiv.org/abs/1905.13319} {Mathqa: Towards interpretable math word problem solving with operation-based formalisms}.

\bibitem[{Azerbayev et~al.(2023)Azerbayev, Schoelkopf, Paster, Santos, McAleer, Jiang, Deng, Biderman, and Welleck}]{azerbayev2023llemma}
Zhangir Azerbayev, Hailey Schoelkopf, Keiran Paster, Marco~Dos Santos, Stephen McAleer, Albert~Q. Jiang, Jia Deng, Stella Biderman, and Sean Welleck. 2023.
\newblock \href {http://arxiv.org/abs/2310.10631} {Llemma: An open language model for mathematics}.

\bibitem[{Bai et~al.(2023)Bai, Bai, Chu, Cui, Dang, Deng, Fan, Ge, Han, Huang et~al.}]{bai2023qwen}
Jinze Bai, Shuai Bai, Yunfei Chu, Zeyu Cui, Kai Dang, Xiaodong Deng, Yang Fan, Wenbin Ge, Yu~Han, Fei Huang, et~al. 2023.
\newblock Qwen technical report.
\newblock \emph{arXiv preprint arXiv:2309.16609}.

\bibitem[{Cobbe et~al.(2021)Cobbe, Kosaraju, Bavarian, Chen, Jun, Kaiser, Plappert, Tworek, Hilton, Nakano, Hesse, and Schulman}]{gsm8k}
Karl Cobbe, Vineet Kosaraju, Mohammad Bavarian, Mark Chen, Heewoo Jun, Lukasz Kaiser, Matthias Plappert, Jerry Tworek, Jacob Hilton, Reiichiro Nakano, Christopher Hesse, and John Schulman. 2021.
\newblock \href {http://arxiv.org/abs/2110.14168} {Training verifiers to solve math word problems}.
\newblock \emph{CoRR}, abs/2110.14168.

\bibitem[{Contributors(2023)}]{2023opencompass}
OpenCompass Contributors. 2023.
\newblock Opencompass: A universal evaluation platform for foundation models.
\newblock \url{https://github.com/open-compass/opencompass}.

\bibitem[{DeepSeek-AI et~al.(2024)DeepSeek-AI, :, Bi, Chen, Chen, Chen, Dai, Deng, Ding, Dong, Du, Fu, Gao, Gao, Gao, Ge, Guan, Guo, Guo, Hao, Hao, He, Hu, Huang, Li, Li, Li, Li, Li, Liang, Lin, Liu, Liu, Liu, Liu, Liu, Liu, Lu, Lu, Luo, Ma, Nie, Pei, Piao, Qiu, Qu, Ren, Ren, Ruan, Sha, Shao, Song, Su, Sun, Sun, Tang, Wang, Wang, Wang, Wang, Wang, Wu, Wu, Xie, Xie, Xie, Xiong, Xu, Xu, Xu, Yang, You, Yu, Yu, Zhang, Zhang, Zhang, Zhang, Zhang, Zhang, Zhang, Zhang, Zhao, Zhao, Zhou, Zhou, Zhu, and Zou}]{deepseek}
DeepSeek-AI, :, Xiao Bi, Deli Chen, Guanting Chen, Shanhuang Chen, Damai Dai, Chengqi Deng, Honghui Ding, Kai Dong, Qiushi Du, Zhe Fu, Huazuo Gao, Kaige Gao, Wenjun Gao, Ruiqi Ge, Kang Guan, Daya Guo, Jianzhong Guo, Guangbo Hao, Zhewen Hao, Ying He, Wenjie Hu, Panpan Huang, Erhang Li, Guowei Li, Jiashi Li, Yao Li, Y.~K. Li, Wenfeng Liang, Fangyun Lin, A.~X. Liu, Bo~Liu, Wen Liu, Xiaodong Liu, Xin Liu, Yiyuan Liu, Haoyu Lu, Shanghao Lu, Fuli Luo, Shirong Ma, Xiaotao Nie, Tian Pei, Yishi Piao, Junjie Qiu, Hui Qu, Tongzheng Ren, Zehui Ren, Chong Ruan, Zhangli Sha, Zhihong Shao, Junxiao Song, Xuecheng Su, Jingxiang Sun, Yaofeng Sun, Minghui Tang, Bingxuan Wang, Peiyi Wang, Shiyu Wang, Yaohui Wang, Yongji Wang, Tong Wu, Y.~Wu, Xin Xie, Zhenda Xie, Ziwei Xie, Yiliang Xiong, Hanwei Xu, R.~X. Xu, Yanhong Xu, Dejian Yang, Yuxiang You, Shuiping Yu, Xingkai Yu, B.~Zhang, Haowei Zhang, Lecong Zhang, Liyue Zhang, Mingchuan Zhang, Minghua Zhang, Wentao Zhang, Yichao Zhang, Chenggang Zhao, Yao Zhao, Shangyan Zhou, Shunfeng
  Zhou, Qihao Zhu, and Yuheng Zou. 2024.
\newblock \href {http://arxiv.org/abs/2401.02954} {Deepseek llm: Scaling open-source language models with longtermism}.

\bibitem[{He et~al.(2022)He, Li, Jin, Wang, Xu, and Lin}]{conghui2022opendatalab}
Conghui He, Wei Li, Zhenjiang Jin, Bin Wang, Chao Xu, and Dahua Lin. 2022.
\newblock Opendatalab: Empowering general artificial intelligence with open datasets.
\newblock \url{https://opendatalab.com}.

\bibitem[{Hendrycks et~al.(2021{\natexlab{a}})Hendrycks, Burns, Basart, Zou, Mazeika, Song, and Steinhardt}]{hendrycks2021measuring}
Dan Hendrycks, Collin Burns, Steven Basart, Andy Zou, Mantas Mazeika, Dawn Song, and Jacob Steinhardt. 2021{\natexlab{a}}.
\newblock \href {http://arxiv.org/abs/2009.03300} {Measuring massive multitask language understanding}.

\bibitem[{Hendrycks et~al.(2021{\natexlab{b}})Hendrycks, Burns, Kadavath, Arora, Basart, Tang, Song, and Steinhardt}]{math}
Dan Hendrycks, Collin Burns, Saurav Kadavath, Akul Arora, Steven Basart, Eric Tang, Dawn Song, and Jacob Steinhardt. 2021{\natexlab{b}}.
\newblock Measuring mathematical problem solving with the math dataset.
\newblock \emph{Cornell University - arXiv,Cornell University - arXiv}.

\bibitem[{Hosseini et~al.(2014)Hosseini, Hajishirzi, Etzioni, and Kushman}]{AddSub}
Mohammad~Javad Hosseini, Hannaneh Hajishirzi, Oren Etzioni, and Nate Kushman. 2014.
\newblock \href {https://doi.org/10.3115/v1/d14-1058} {Learning to solve arithmetic word problems with verb categorization}.
\newblock In \emph{Proceedings of the 2014 Conference on Empirical Methods in Natural Language Processing (EMNLP)}.

\bibitem[{Huang et~al.(2023)Huang, Bai, Zhu, Zhang, Zhang, Su, Liu, Lv, Zhang, Lei, Fu, Sun, and He}]{huang2023ceval}
Yuzhen Huang, Yuzhuo Bai, Zhihao Zhu, Junlei Zhang, Jinghan Zhang, Tangjun Su, Junteng Liu, Chuancheng Lv, Yikai Zhang, Jiayi Lei, Yao Fu, Maosong Sun, and Junxian He. 2023.
\newblock \href {http://arxiv.org/abs/2305.08322} {C-eval: A multi-level multi-discipline chinese evaluation suite for foundation models}.

\bibitem[{Jiang et~al.(2024)Jiang, Sablayrolles, Roux, Mensch, Savary, Bamford, Chaplot, de~las Casas, Hanna, Bressand, Lengyel, Bour, Lample, Lavaud, Saulnier, Lachaux, Stock, Subramanian, Yang, Antoniak, Scao, Gervet, Lavril, Wang, Lacroix, and Sayed}]{jiang2024mixtral}
Albert~Q. Jiang, Alexandre Sablayrolles, Antoine Roux, Arthur Mensch, Blanche Savary, Chris Bamford, Devendra~Singh Chaplot, Diego de~las Casas, Emma~Bou Hanna, Florian Bressand, Gianna Lengyel, Guillaume Bour, Guillaume Lample, Lélio~Renard Lavaud, Lucile Saulnier, Marie-Anne Lachaux, Pierre Stock, Sandeep Subramanian, Sophia Yang, Szymon Antoniak, Teven~Le Scao, Théophile Gervet, Thibaut Lavril, Thomas Wang, Timothée Lacroix, and William~El Sayed. 2024.
\newblock \href {http://arxiv.org/abs/2401.04088} {Mixtral of experts}.

\bibitem[{Kaplan et~al.(2020)Kaplan, McCandlish, Henighan, Brown, Chess, Child, Gray, Radford, Wu, and Amodei}]{kaplan2020scaling}
Jared Kaplan, Sam McCandlish, Tom Henighan, Tom~B. Brown, Benjamin Chess, Rewon Child, Scott Gray, Alec Radford, Jeffrey Wu, and Dario Amodei. 2020.
\newblock \href {http://arxiv.org/abs/2001.08361} {Scaling laws for neural language models}.

\bibitem[{Koncel-Kedziorski et~al.(2015)Koncel-Kedziorski, Hajishirzi, Sabharwal, Etzioni, and Ang}]{SingleEq}
Rik Koncel-Kedziorski, Hannaneh Hajishirzi, Ashish Sabharwal, Oren Etzioni, and Siena~Dumas Ang. 2015.
\newblock Parsing algebraic word problems into equations.
\newblock \emph{Transactions of the Association for Computational Linguistics}, 3:585--597.

\bibitem[{Kushman et~al.(2014)Kushman, Artzi, Zettlemoyer, and Barzilay}]{Alg514}
Nate Kushman, Yoav Artzi, Luke Zettlemoyer, and Regina Barzilay. 2014.
\newblock Learning to automatically solve algebra word problems.
\newblock In \emph{Proceedings of the 52nd Annual Meeting of the Association for Computational Linguistics (Volume 1: Long Papers)}, pages 271--281.

\bibitem[{Ling et~al.(2017)Ling, Yogatama, Dyer, and Blunsom}]{AQuA}
Wang Ling, Dani Yogatama, Chris Dyer, and Phil Blunsom. 2017.
\newblock Program induction by rationale generation: Learning to solve and explain algebraic word problems.
\newblock \emph{arXiv preprint arXiv:1705.04146}.

\bibitem[{Liu et~al.(2023{\natexlab{a}})Liu, Hu, Zhou, Ding, Li, Zeng, He, Chen, Jiang, Zhou, and He}]{mathematic_survey}
Wentao Liu, Hanglei Hu, Jie Zhou, Yuyang Ding, Junsong Li, Jiayi Zeng, Mengliang He, Qin Chen, Bo~Jiang, Aimin Zhou, and Liang He. 2023{\natexlab{a}}.
\newblock Mathematical language models: A survey.

\bibitem[{Liu et~al.(2023{\natexlab{b}})Liu, Duan, Zhang, Li, Zhang, Zhao, Yuan, Wang, He, Liu, Chen, and Lin}]{mmbench}
Yuan Liu, Haodong Duan, Yuanhan Zhang, Bo~Li, Songyang Zhang, Wangbo Zhao, Yike Yuan, Jiaqi Wang, Conghui He, Ziwei Liu, Kai Chen, and Dahua Lin. 2023{\natexlab{b}}.
\newblock \href {http://arxiv.org/abs/2307.06281} {Mmbench: Is your multi-modal model an all-around player?}

\bibitem[{Miao et~al.(2021)Miao, Liang, and Su}]{ASDiv}
Shen-Yun Miao, Chao-Chun Liang, and Keh-Yih Su. 2021.
\newblock \href {http://arxiv.org/abs/2106.15772} {A diverse corpus for evaluating and developing english math word problem solvers}.

\bibitem[{Mishra et~al.(2023)Mishra, Finlayson, Lu, Tang, Welleck, Baral, Rajpurohit, Tafjord, Sabharwal, Clark, and Kalyan}]{Lila}
Swaroop Mishra, Matthew Finlayson, Pan Lu, Leonard Tang, Sean Welleck, Chitta Baral, Tanmay Rajpurohit, Oyvind Tafjord, Ashish Sabharwal, Peter Clark, and Ashwin Kalyan. 2023.
\newblock \href {http://arxiv.org/abs/2210.17517} {Lila: A unified benchmark for mathematical reasoning}.

\bibitem[{Mishra et~al.(2022)Mishra, Mitra, Varshney, Sachdeva, Clark, Baral, and Kalyan}]{NumGLUE}
Swaroop Mishra, Arindam Mitra, Neeraj Varshney, Bhavdeep Sachdeva, Peter Clark, Chitta Baral, and Ashwin Kalyan. 2022.
\newblock \href {http://arxiv.org/abs/2204.05660} {Numglue: A suite of fundamental yet challenging mathematical reasoning tasks}.

\bibitem[{Roy and Roth(2016)}]{MultiArith}
Subhro Roy and Dan Roth. 2016.
\newblock \href {http://arxiv.org/abs/1608.01413} {Solving general arithmetic word problems}.

\bibitem[{Roy et~al.(2015)Roy, Vieira, and Roth}]{SingleOp}
Subhro Roy, Tim Vieira, and Dan Roth. 2015.
\newblock \href {https://doi.org/10.1162/tacl_a_00118} {{Reasoning about Quantities in Natural Language}}.
\newblock \emph{Transactions of the Association for Computational Linguistics}, 3:1--13.

\bibitem[{Saxton et~al.(2019)Saxton, Grefenstette, Hill, and Kohli}]{Mathematics_Dataset}
David Saxton, Edward Grefenstette, Felix Hill, and Pushmeet Kohli. 2019.
\newblock \href {http://arxiv.org/abs/1904.01557} {Analysing mathematical reasoning abilities of neural models}.

\bibitem[{Shao et~al.(2024)Shao, Wang, Zhu, Xu, Song, Zhang, Li, Wu, and Guo}]{deepseekmath}
Zhihong Shao, Peiyi Wang, Qihao Zhu, Runxin Xu, Junxiao Song, Mingchuan Zhang, Y.~K. Li, Y.~Wu, and Daya Guo. 2024.
\newblock \href {http://arxiv.org/abs/2402.03300} {Deepseekmath: Pushing the limits of mathematical reasoning in open language models}.

\bibitem[{Team(2023{\natexlab{a}})}]{team2023internlm}
InternLM Team. 2023{\natexlab{a}}.
\newblock Internlm: A multilingual language model with progressively enhanced capabilities.

\bibitem[{Team(2023{\natexlab{b}})}]{lagent2023}
Lagent~Developer Team. 2023{\natexlab{b}}.
\newblock {Lagent: InternLM} a lightweight open-source framework that allows users to efficiently build large language model(llm)-based agents.
\newblock \url{https://github.com/InternLM/lagent}.

\bibitem[{Touvron et~al.(2023)Touvron, Martin, Stone, Albert, Almahairi, Babaei, Bashlykov, Batra, Bhargava, Bhosale et~al.}]{touvron2023llama2}
Hugo Touvron, Louis Martin, Kevin Stone, Peter Albert, Amjad Almahairi, Yasmine Babaei, Nikolay Bashlykov, Soumya Batra, Prajjwal Bhargava, Shruti Bhosale, et~al. 2023.
\newblock Llama 2: Open foundation and fine-tuned chat models.
\newblock \emph{arXiv preprint arXiv:2307.09288}.

\bibitem[{Upadhyay and Chang(2017)}]{draw}
Shyam Upadhyay and Ming-Wei Chang. 2017.
\newblock \href {http://arxiv.org/abs/1609.07197} {Annotating derivations: A new evaluation strategy and dataset for algebra word problems}.

\bibitem[{Wang et~al.(2024)Wang, Hu, Lu, Zhu, Zhang, Subramaniam, Loomba, Zhang, Sun, and Wang}]{wang2024scibench}
Xiaoxuan Wang, Ziniu Hu, Pan Lu, Yanqiao Zhu, Jieyu Zhang, Satyen Subramaniam, Arjun~R. Loomba, Shichang Zhang, Yizhou Sun, and Wei Wang. 2024.
\newblock \href {http://arxiv.org/abs/2307.10635} {Scibench: Evaluating college-level scientific problem-solving abilities of large language models}.

\bibitem[{Wang et~al.(2017)Wang, Liu, and Shi}]{Math23k}
Yan Wang, Xiaojiang Liu, and Shuming Shi. 2017.
\newblock \href {https://doi.org/10.18653/v1/D17-1088} {Deep neural solver for math word problems}.
\newblock In \emph{Proceedings of the 2017 Conference on Empirical Methods in Natural Language Processing}, pages 845--854, Copenhagen, Denmark. Association for Computational Linguistics.

\bibitem[{Yang et~al.(2023)Yang, Xiao, Wang, Zhang, Yin, Lv, Pan, Wang, Yan, Yang et~al.}]{yang2023baichuan}
Aiyuan Yang, Bin Xiao, Bingning Wang, Borong Zhang, Chao Yin, Chenxu Lv, Da~Pan, Dian Wang, Dong Yan, Fan Yang, et~al. 2023.
\newblock Baichuan 2: Open large-scale language models.
\newblock \emph{arXiv preprint arXiv:2309.10305}.

\bibitem[{Yao et~al.(2023)Yao, Zhao, Yu, Du, Shafran, Narasimhan, and Cao}]{yao2023react}
Shunyu Yao, Jeffrey Zhao, Dian Yu, Nan Du, Izhak Shafran, Karthik Narasimhan, and Yuan Cao. 2023.
\newblock \href {http://arxiv.org/abs/2210.03629} {React: Synergizing reasoning and acting in language models}.

\bibitem[{Ying et~al.(2024)Ying, Zhang, Li, Zhou, Shao, Fei, Ma, Hong, Liu, Wang, Wang, Wu, Li, Zhou, Liu, Zhang, Zhang, Yan, Qiu, Wang, Chen, and Lin}]{ying2024internlmmath}
Huaiyuan Ying, Shuo Zhang, Linyang Li, Zhejian Zhou, Yunfan Shao, Zhaoye Fei, Yichuan Ma, Jiawei Hong, Kuikun Liu, Ziyi Wang, Yudong Wang, Zijian Wu, Shuaibin Li, Fengzhe Zhou, Hongwei Liu, Songyang Zhang, Wenwei Zhang, Hang Yan, Xipeng Qiu, Jiayu Wang, Kai Chen, and Dahua Lin. 2024.
\newblock \href {http://arxiv.org/abs/2402.06332} {Internlm-math: Open math large language models toward verifiable reasoning}.

\bibitem[{Yu et~al.(2023)Yu, Jiang, Shi, Yu, Liu, Zhang, Kwok, Li, Weller, and Liu}]{MetaMath}
Longhui Yu, Weisen Jiang, Han Shi, Jincheng Yu, Zhengying Liu, Yu~Zhang, James~T. Kwok, Zhenguo Li, Adrian Weller, and Weiyang Liu. 2023.
\newblock \href {http://arxiv.org/abs/2309.12284} {Metamath: Bootstrap your own mathematical questions for large language models}.

\bibitem[{Yuan et~al.(2023)Yuan, Yuan, Tan, Wang, and Huang}]{math401}
Zheng Yuan, Hongyi Yuan, Chuanqi Tan, Wei Wang, and Songfang Huang. 2023.
\newblock How well do large language models perform in arithmetic tasks?

\bibitem[{Yue et~al.(2023)Yue, Qu, Zhang, Fu, Huang, Sun, Su, and Chen}]{mammoth}
Xiang Yue, Xingwei Qu, Ge~Zhang, Yao Fu, Wenhao Huang, Huan Sun, Yu~Su, and Wenhu Chen. 2023.
\newblock \href {http://arxiv.org/abs/2309.05653} {Mammoth: Building math generalist models through hybrid instruction tuning}.

\bibitem[{Zeng et~al.(2022)Zeng, Liu, Du, Wang, Lai, Ding, Yang, Xu, Zheng, Xia et~al.}]{zeng2022glm}
Aohan Zeng, Xiao Liu, Zhengxiao Du, Zihan Wang, Hanyu Lai, Ming Ding, Zhuoyi Yang, Yifan Xu, Wendi Zheng, Xiao Xia, et~al. 2022.
\newblock Glm-130b: An open bilingual pre-trained model.
\newblock \emph{arXiv preprint arXiv:2210.02414}.

\end{thebibliography}

\appendix

\definecolor{lightgray}{gray}{0.95}
\lstdefinestyle{prompt}{
    basicstyle=\ttfamily\fontsize{7pt}{8pt}\selectfont,
    frame=none,
    breaklines=true,
    backgroundcolor=\color{lightgray},
    breakatwhitespace=true,
    breakindent=0pt,
    escapeinside={(*@}{@*)},
    numbers=none,
    numbersep=5pt,
    xleftmargin=5pt,
}
\tcbset{
  aibox/.style={
    top=10pt,
    colback=white,
    colframe=black,
    colbacktitle=black,
    enhanced,
    center,
    attach boxed title to top left={yshift=-0.1in,xshift=0.15in},
    boxed title style={boxrule=0pt,colframe=white,},
  }
}

\section{MathBench Statistics}

\subsection{Dataset Statistics}
\label{app:data_Statistics}
The detailed statistics of MathBench questions, \Cref{tab:mathbench_app_theory} for the data distribution of theoretical and application questions across various stages, \Cref{tab:Statistics with subject area} for fine-grained knowledge levels.

\begin{table}[!th]
\centering
\caption{Detailed Composition of the MathBench}
\label{tab:mathbench_app_theory}
\resizebox{\columnwidth}{!}{
\begin{tabular}{lccc|c}
\toprule
\textbf{Stage} & \textbf{Type} & \textbf{English} & \textbf{Chinese} & \textbf{Total} \\
\midrule
\multirow{2}{*}{Arithmetic} & Theoretical & - & - & - \\
                             & Application & 300 & - & 300 \\
\midrule
\multirow{2}{*}{Primary} & Theoretical & 109 & 208 & 317 \\
                         & Application & 150 & 150 & 300 \\
\midrule
\multirow{2}{*}{Middle} & Theoretical & 175 & 334 & 509 \\
                        & Application & 150 & 150 & 300 \\
\midrule
\multirow{2}{*}{High} & Theoretical & 281 & 470 & 751 \\
                      & Application & 150 & 150 & 300 \\
\midrule
\multirow{2}{*}{College} & Theoretical & 316 & 316 & 632 \\
                         & Application & 150 & 150 & 300 \\
\bottomrule
\end{tabular}}
\end{table}

\begin{table*}[!th]
\centering
\caption{\textbf{MathBench Subject Area Statistics}. Data is shown at the Subject Area level for conciseness, omitting the more detailed Topic level due to its breadth.}
\begin{tabular}{llcc|c}
\toprule
\textbf{Stage}                     & \textbf{Subject Area}                   & \textbf{English} & \textbf{Chinese} & \textbf{Total} \\ 
\midrule
\multirow{3}{*}{Primary \& Arith.} & Introduction to Numbers and Algebra                     & 44               & 73               & 117            \\
                                   & Introduction to Geometry                & 10               & 62               & 72             \\
                                   & Comprehensive Application               & 55               & 73               & 128            \\
\midrule
\multirow{3}{*}{Middle}            & Basic Numbers and Algebra               & 133              & 182              & 315            \\
                                   & Basic Geometry                          & 33               & 137              & 170            \\
                                   & Basic Probability and Statistics        & 9                & 15               & 24             \\
\midrule
\multirow{3}{*}{High}              & Intermediate Numbers and Algebra        & 146              & 189              & 335            \\
                                   & Intermediate Geometry                   & 114              & 219              & 333            \\
                                   & Intermediate Probability and Statistics & 21               & 62               & 83             \\
\midrule
\multirow{3}{*}{College}           & Advanced Mathematics                    & 119              & 119              & 238            \\
                                   & Linear Algebra                          & 99               & 99               & 198            \\
                                   & Probability and Statistics              & 98               & 98               & 196            \\
\bottomrule
\label{tab:Statistics with subject area}
\end{tabular}
\end{table*}

\subsection{Data collection details}
For self-collected questions in MathBench, We primarily collect through the following methods:

For the Primary stage GSM-X-CN and GSM-X-Plus datasets, we semi-automatically generate new questions using GPT-4. 
Specifically, the construction of the GSM-X-CN Chinese question set involved two steps:

We first translate English questions in GSM8k test set into Chinese using GPT-4, resulting in a Chinese version of GSM8k. 
We then replace the entity names under the Chinese context while ensuring that the questions' meanings remained unchanged. 
This process creates elementary-level questions suitable for Chinese Q\&A.

For the GSM-X-Plus dataset, which is in English, 
we generate new questions by first generating solution code for the original test set questions. 
We then replace some numeric parameters (taken from the original questions) in the question with multiples of the variable $k$. 
By executing the modified solution code, we obtain the new answers. 
In MathBench, we set $k \in (2, 10)$.

For exams such as AMC, GaoKao, ZhongKao, \textit{etc.}, we initially collect relevant questions from the Internet. 
These questions are then underwent processing and annotation by domain experts.
Questions for primary and secondary education levels are handled and annotated by undergraduate students, 
while questions for university-level exams were processed and annotated by graduate students specializing in mathematics or computer science. 
The description of knowledge-based questions is provided in \Cref{par: Theoretical Knowledge Questions}.

In addition to the self-collected datasets described above, we also incorporate questions from the following open-source datasets: CEVAL~\cite{huang2023ceval}, MMLU~\citep{hendrycks2021measuring}, Arithmetic-HG, Math401~\citep{math401} and SciBench~\citep{wang2024scibench}. We download the data from OpenDataLab~\citep{conghui2022opendatalab}. All open-source datasets we used are MIT License.

\subsection{Quality Screening}
\label{app: quality_screening}
Given the wide variety of sources and types of questions,
we notice that the following issues may affect the benchmark quality: 
1. Intrinsic errors in the questions, such as being unanswerable or having multiple correct answers.
2. Questions of low evaluation value, too difficult or too trivial for the intended education stage.

All of the above situations can easily lead to unstable model responses and increased probability of incorrect answers in CircularEval. 
To address these issues, we employ a novel semi-automated question filtering approach for quality screening. 

Specifically, we use GPT-4 to perform Circular Evaluation (CE) on all questions. 
We then select questions that GPT-4 answered incorrectly 0, 1, or 2 times out of four attempts ($CE-0$, $CE-1$, $CE-2$) for manual review to ensure the overall question quality.

\section{Detailed Experimental Results}

\subsection{Overall Results}
The overall experimental results for chat models are shown in \Cref{tab:main_results}, and \Cref{tab:base_results} for base models.
We report the average score of theoretical and application questions for stages except \textit{Arithmetic}, which is predominantly indicative of pure computational prowess.

\begin{table*}[!thb]
\centering
\caption{\textbf{Overall Comparison of Chat Models on MathBench}. Models are classified into three categories according to their purpose and origin. The model name in \textbf{bold} indicates the top performer among Open-source or Closed-source models, while an \underline{underline} signifies the leading model within a similar parameter size group.}
\vspace{-2mm}
\label{tab:main_results}
\resizebox{\linewidth}{!}{
\tablestyle{10pt}{1.2}
\begin{tabular}{lccccc|c}
\hline
\textbf{Models} & \textbf{Arith} & \textbf{Primary} & \textbf{Middle} & \textbf{High} & \textbf{College} & \textbf{Avg.} \\
\hline
\multicolumn{7}{c}{\purplecolor{\textit{\ding{72}Closed-source Models}}} \\
GPT-3.5-Turbo-0125 & 72.7 & 71.2 & 42.0 & 32.8 & 33.4 & 48.8 \\
GLM4 & 61.7 & 84.3 & 67.6 & 51.2 & 40.6 & 62.2 \\
GPT-4-0125-Preview & 76.0 & 84.8 & 70.0 & 56.7 & 54.3 & 68.6 \\
Qwen-Max-0428 & 72.3 & 88.4 & 74.1 & 59.2 & 51.1 & 69.8 \\
DeepSeek-V2-API & 82.7 & 89.1 & 71.4 & 54.8 & 52.8 & 69.9 \\
Claude-3-Opus & \textbf{85.7} & 85.5 & 68.5 & 57.6 & 60.5 & 70.9 \\
GPT-4o-2024-05-13 & 77.7 & \textbf{89.9} & \textbf{82.3} & \textbf{70.5} & \textbf{69.8} & \textbf{79.0} \\
\hline
\multicolumn{7}{c}{\bluecolor{\textit{$\heartsuit$Open-source Chat Models}}} \\
DeepSeek-7B-Chat & 48.3 & 40.5 & 17.4 & 9.4 & 8.1 & 22.1 \\
ChatGLM3-6B & 38.0 & 41.3 & 23.0 & 12.8 & 6.8 & 23.2 \\
Yi-6B-Chat & 35.3 & 42.2 & 20.3 & 12.4 & 14.1 & 24.5 \\
Qwen-7B-Chat & 50.7 & 51.9 & 32.7 & 21.1 & 18.6 & 33.9 \\
InternLM2-Chat-7B & \underline{52.0} & \underline{66.8} & \underline{42.9} & \underline{29.5} & \underline{25.7} & \underline{43.5} \\
\hline
Baichuan2-13B-Chat & 40.0 & 45.0 & 25.3 & 14.4 & 11.4 & 26.4 \\
Yi-34B-Chat & 50.7 & 66.5 & 40.0 & 29.1 & 27.3 & 43.1 \\
Llama-3-8B-Instruct & 54.7 & 65.6 & 38.1 & 31.2 & \underline{33.8} & 44.4 \\
InternLM2-Chat-20B & 62.3 & \underline{68.6} & 46.9 & 37.3 & 28.1 & 47.8 \\
Qwen-14B-Chat & \underline{63.7} & 66.6 & \underline{51.5} & \underline{35.4} & 30.7 & \underline{49.1} \\
\hline
DeepSeek-67B-Chat & 62.0 & 75.4 & 49.5 & 38.5 & 38.3 & 53.1 \\
Llama-3-70B-Instruct & 70.3 & 78.7 & 58.7 & 50.4 & 52.6 & 61.8 \\
Qwen-72B-Chat & \textbf{72.0} & 81.3 & 67.3 & 49.6 & 44.4 & 63.4 \\
Qwen1.5-110B-Chat & 70.3 & \textbf{87.9} & \textbf{74.5} & \textbf{61.9} & \textbf{54.7} & \textbf{71.2} \\
\hline
\multicolumn{7}{c}{\greencolor{\textit{$\triangle$Mathematical Models}}} \\
MammoTH-7B & 27.0 & 17.9 & 5.9 & 5.0 & 3.5 & 10.1 \\
MammoTH-13B & 35.0 & 35.2 & 11.8 & 9.9 & 11.0 & 19.0 \\
Metamath-Llemma-7B & 51.7 & 43.8 & 20.9 & 18.6 & 15.5 & 28.0 \\
MammoTH-70B & 35.7 & 59.1 & 29.1 & 25.0 & 25.3 & 36.0 \\
InternLM2-Chat-Math-7B & 53.7 & 66.3 & 50.8 & 35.0 & 27.3 & 46.8 \\
DeepSeek-Math-7B-Instruct & 61.0 & 73.7 & 44.4 & 37.0 & 32.3 & 49.3 \\
InternLM2-Chat-Math-20B & 58.7 & 71.6 & 57.1 & 42.6 & 32.8 & 53.1 \\
DeepSeek-Math-7B-RL & \textbf{68.0} & \textbf{81.5} & \textbf{58.2} & \textbf{47.2} & \textbf{45.8} & \textbf{60.4} \\
\hline
\end{tabular}}
\end{table*}

\begin{table*}[!thb]
\centering
\caption{\textbf{Overall Comparison of Base Models on MathBench}. Models are classified into categories based on their parameter size and the dataset they were trained on. The model name in \textbf{bold} indicates the top performer within all base models, while an \underline{underline} signifies the leading model within a similar parameter size group.}
\vspace{-2mm}
\label{tab:base_results}
\resizebox{\linewidth}{!}{
\tablestyle{10pt}{1.2}
\begin{tabular}{lccccc|c}
\hline
\textbf{Models} & \textbf{Arithmetic} & \textbf{Primary} & \textbf{Middle} & \textbf{High} & \textbf{College} & \textbf{Average} \\
\hline
\multicolumn{7}{c}{\textit{\bluecolor{$\heartsuit$Open-source Base Models}}} \\
Llama-2-7B & 28.0 & 11.3 & 16.1 & 19.1 & 20.0 & 18.9 \\
Deepseek-7B-Base & 31.0 & 19.1 & 22.2 & 22.5 & 22.9 & 23.5 \\
Baichuan2-7B-Base & 44.0 & 29.8 & 30.8 & 26.9 & 27.2 & 31.7 \\
Qwen-7B & 44.3 & 42.2 & 37.4 & 28.3 & 31.2 & 36.7 \\
ChatGLM3-6B-Base & 39.7 & 50.7 & 46.1 & 37.5 & 35.3 & 41.9 \\
InternLM2-7B & \underline{49.0} & \underline{56.1} & \underline{46.3} & \underline{41.2} & \underline{44.2} & \underline{47.4} \\
\hline
Llama-2-13B & 30.0 & 25.4 & 26.1 & 24.7 & 25.6 & 26.4 \\
Baichuan2-13B-Base & 47.7 & 44.8 & 39.3 & 30.1 & 39.3 & 40.2 \\
InternLM2-20B & \underline{57.3} & 61.8 & 47.0 & 42.7 & 46.0 & 51.2 \\
Qwen-14B & 52.0 & \underline{63.0} & \underline{57.4} & \underline{45.8} & \underline{49.7} & \underline{53.6} \\
\hline
Llama-2-70B & 44.3 & 49.2 & 39.1 & 34.8 & 46.8 & 42.9 \\
Mixtral-8x7B-v0.1 & 55.3 & 52.6 & 42.1 & 40.1 & 51.9 & 48.4 \\
Deepseek-67B-Base & 45.3 & 64.2 & 51.6 & 44.0 & 50.7 & 51.2 \\
Qwen-72B & \textbf{62.3} & \textbf{77.9} & \textbf{69.8} & \textbf{64.4} & \textbf{64.4} & \textbf{67.8} \\
\multicolumn{7}{c}{\textit{\greencolor{$\triangle$Mathematical Models}}} \\
Llemma-7B & 41.3 & 25.8 & 30.7 & 32.0 & 38.4 & 33.6 \\
InternLM2-Base-Math-7B & 46.0 & 38.2 & 40.4 & 34.7 & 43.2 & 40.6 \\
Llemma-34B & 44.3 & 43.8 & 40.4 & 37.4 & 46.6 & 42.5 \\
InternLM2-Base-Math-20B & 48.0 & 49.5 & 47.3 & 44.0 & 46.8 & 47.1 \\
Deepseek-Math-7B-Base & \underline{58.3} & \underline{62.3} & \underline{55.7} & \underline{50.6} & \underline{57.5} & \underline{56.9} \\
\hline
\end{tabular}}
\end{table*}

\subsection{\textsc{Results with Accuracy}}
\begin{table*}[!thb]
\centering
\caption{\textbf{Overall Comparison with Accuracy on MathBench}. Models are classified into three categories according to their purpose and origin.}
\vspace{-2mm}
\label{tab:acc_results}
\resizebox{\linewidth}{!}{
\begin{tabular}{lccccc|c}
\hline
\textbf{Models} & \textbf{Arithmetic} & \textbf{Primary} & \textbf{Middle} & \textbf{High} & \textbf{College} & \textbf{Average} \\
\hline
\multicolumn{7}{c}{\purplecolor{\textit{\ding{72}Closed-source Models}}} \\
GPT-3.5-Turbo-0125 & 74.8 & 71.8 & 79.5 & 85.3 & 85.3 & 77.9 \\
Claude-3-Opus & 83.8 & 85.2 & 90.0 & 87.2 & 86.0 & 86.4 \\
DeepSeek-V2-API & 83.9 & 82.0 & 89.1 & 90.5 & 89.7 & 87.0 \\
GLM4 & 83.9 & 81.3 & 92.2 & 91.5 & 93.7 & 88.5 \\
GPT-4-0125-Preview & 91.1 & 86.4 & 91.4 & 92.4 & 92.7 & 90.3 \\
Qwen-Max-0428 & 88.0 & 87.5 & 92.9 & 94.8 & 87.5 & 90.8 \\
GPT-4o-2024-05-13 & 91.9 & 91.5 & 93.7 & 96.2 & 94.0 & 93.3 \\
\hline
\multicolumn{7}{c}{\bluecolor{\textit{$\heartsuit$Open-source Chat Models}}} \\
DeepSeek-7B-Chat & 48.3 & 40.5 & 17.4 & 9.4 & 8.1 & 22.1 \\
Yi-6B-Chat & 35.7 & 41.3 & 23.0 & 12.8 & 6.8 & 23.2 \\
Qwen-7B-Chat & 50.7 & 51.9 & 32.7 & 21.1 & 18.6 & 33.9 \\
InternLM2-Chat-7B & 52.0 & 66.8 & 42.9 & 29.5 & 25.7 & 43.5 \\
ChatGLM3-6B & 41.0 & 53.2 & 51.2 & 38.9 & 34.0 & 43.7 \\
Llama-3-8B-Instruct & 54.7 & 65.6 & 38.1 & 31.2 & 33.8 & 44.4 \\
\hline
Baichuan2-13B-Chat & 40.0 & 45.0 & 25.3 & 14.4 & 11.4 & 26.4 \\
Yi-34B-Chat & 50.7 & 66.5 & 40.0 & 29.1 & 27.3 & 43.1 \\
InternLM2-Chat-20B & 62.3 & 68.6 & 46.9 & 37.3 & 28.1 & 47.8 \\
Qwen-14B-Chat & 63.7 & 66.6 & 51.5 & 35.4 & 30.7 & 49.1 \\
\hline
DeepSeek-67B-Chat & 62.0 & 75.4 & 49.5 & 38.5 & 38.3 & 53.1 \\
Llama-3-70B-Instruct & 70.3 & 78.7 & 58.7 & 50.4 & 52.6 & 61.8 \\
Qwen-72B-Chat & 72.0 & 81.3 & 67.3 & 49.6 & 44.4 & 63.4 \\
Qwen1.5-110B-Chat & 70.3 & 87.9 & 74.5 & 61.9 & 54.7 & 71.2 \\
\hline
\multicolumn{7}{c}{\greencolor{\textit{$\triangle$Mathematical Models}}} \\
MammoTH-7B & 27.0 & 17.9 & 5.9 & 5.0 & 3.5 & 10.1 \\
MammoTH-13B & 35.0 & 35.2 & 11.8 & 9.9 & 11.0 & 19.0 \\
Metamath-Llemma-7B & 51.7 & 43.8 & 20.9 & 18.6 & 15.5 & 28.0 \\
MammoTH-70B & 35.7 & 59.1 & 29.1 & 25.0 & 25.3 & 36.0 \\
InternLM2-Chat-Math-7B & 53.7 & 66.3 & 50.8 & 35.0 & 27.3 & 46.8 \\
DeepSeek-Math-7B-Instruct & 61.0 & 73.7 & 44.4 & 37.0 & 32.3 & 49.3 \\
InternLM2-Chat-Math-20B & 58.7 & 71.6 & 57.1 & 42.6 & 32.8 & 53.1 \\
DeepSeek-Math-7B-RL & 68.0 & 81.5 & 58.2 & 47.2 & 45.8 & 60.4 \\
\hline
\end{tabular}
}
\end{table*}

The detailed accuracy results are presented in \Cref{tab:acc_results}. 

\subsection{\textsc{Bilingual}}
\label{app: bilingual results}
\begin{table*}[!htb]
    \begin{minipage}{.5\linewidth}
      \centering
        \resizebox{\linewidth}{!}{
        \begin{tabular}{lcccc|c}
        \hline
        \textbf{Models} & \textbf{Primary} & \textbf{Middle} & \textbf{High} & \textbf{College} & \textbf{Avg.} \\
        \hline
        \multicolumn{6}{c}{\purplecolor{\textit{\ding{72}Closed-source Models}}} \\
        GPT-3.5-Turbo-0125 & 77.6 & 43.1 & 44.1 & 41.3 & 51.5 \\
        GLM4 & 82.7 & 59.6 & 60.3 & 47.8 & 62.6 \\
        Qwen-Max-0428 & 85.2 & 63.3 & 60.9 & 50.1 & 64.9 \\
        GPT-4-0125-Preview & 82.4 & 66.7 & 64.6 & 58.2 & 68.0 \\
        DeepSeek-V2-API & 90.7 & 66.8 & 61.7 & 60.9 & 70.0 \\
        Claude-3-Opus & 85.1 & 67.4 & 65.5 & 70.5 & 72.1 \\
        GPT-4o-2024-05-13 & 88.3 & 80.2 & 78.4 & 75.9 & 80.7 \\
        \hline
        \multicolumn{6}{c}{\bluecolor{\textit{$\heartsuit$Open-source Chat Models}}} \\
        DeepSeek-7B-Chat & 41.9 & 16.2 & 9.8 & 8.0 & 19.0 \\
        ChatGLM3-6B & 45.3 & 20.1 & 17.7 & 7.5 & 22.6 \\
        Yi-6B-Chat & 44.5 & 17.7 & 15.6 & 16.3 & 23.5 \\
        Qwen-7B-Chat & 49.4 & 24.6 & 23.3 & 21.6 & 29.7 \\
        InternLM2-Chat-7B & 68.5 & 34.2 & 35.7 & 32.1 & 42.6 \\
        Llama-3-8B-Instruct & 76.4 & 43.8 & 40.5 & 41.3 & 50.5 \\
        \hline
        Baichuan2-13B-Chat & 49.2 & 20.4 & 16.0 & 11.4 & 24.3 \\
        Yi-34B-Chat & 63.5 & 31.1 & 32.8 & 27.6 & 38.8 \\
        Qwen-14B-Chat & 63.3 & 38.7 & 37.0 & 32.6 & 42.9 \\
        InternLM2-Chat-20B & 75.5 & 44.9 & 47.2 & 36.4 & 50.9 \\
        \hline
        DeepSeek-67B-Chat & 77.0 & 48.6 & 47.3 & 41.3 & 53.6 \\
        Qwen-72B-Chat & 80.1 & 53.6 & 49.0 & 45.0 & 56.9 \\
        Qwen-1.5-110B-Chat & 85.0 & 65.0 & 65.5 & 55.8 & 67.9 \\
        Llama-3-70B-Instruct & 88.1 & 68.1 & 64.7 & 62.1 & 70.7 \\
        \hline
        \multicolumn{6}{c}{\greencolor{\textit{$\triangle$Mathematical Models}}} \\
        MammoTH-7B & 26.3 & 9.1 & 8.5 & 6.3 & 12.5 \\
        MammoTH-13B & 49.2 & 18.8 & 15.1 & 16.7 & 24.9 \\
        MetaMath-Llemma-7B & 62.7 & 30.3 & 29.6 & 22.2 & 36.2 \\
        DeepSeek-Math-7B-Instruct & 71.7 & 34.4 & 33.0 & 29.2 & 42.3 \\
        InternLM2-Chat-Math-7B & 66.6 & 32.1 & 31.0 & 28.4 & 39.5 \\
        MammoTH-70B & 70.2 & 31.6 & 30.4 & 31.4 & 40.9 \\
        InternLM2-Chat-Math-20B & 70.7 & 38.3 & 36.3 & 31.8 & 44.3 \\
        DeepSeek-Math-7B-RL & 82.7 & 53.1 & 50.7 & 49.1 & 58.9 \\
        \hline
        \end{tabular}
        }
    \caption*{\small{English Part.}}
    \end{minipage}%
    \hfill
    \begin{minipage}{.5\linewidth}
      \centering
        \resizebox{\linewidth}{!}{
\begin{tabular}{lcccc|c}
\hline
\textbf{Models} & \textbf{Primary} & \textbf{Middle} & \textbf{High} & \textbf{College} & \textbf{Avg.} \\
\hline
\multicolumn{6}{c}{\purplecolor{\textit{\ding{72}Closed-source Models}}} \\
GPT-3.5-Turbo-0125 & 64.8 & 41.0 & 21.6 & 25.5 & 38.2 \\
GLM4 & 86.0 & 75.6 & 42.1 & 33.5 & 59.3 \\
GPT-4-0125-Preview & 87.2 & 73.3 & 48.7 & 50.4 & 64.9 \\
Claude-3-Opus & 86.0 & 69.6 & 49.7 & 50.5 & 63.9 \\
DeepSeek-V2-API & 87.6 & 75.9 & 47.9 & 44.7 & 64.0 \\
Qwen-Max-0428 & 91.6 & 84.9 & 57.4 & 52.1 & 71.5 \\
GPT-4o-2024-05-13 & 91.6 & 84.4 & 62.6 & 63.7 & 75.6 \\
\hline
\multicolumn{6}{c}{\bluecolor{\textit{$\heartsuit$Open-source Chat Models}}} \\
DeepSeek-7B-Chat & 39.0 & 18.5 & 8.9 & 8.3 & 18.7 \\
ChatGLM3-6B & 37.3 & 25.9 & 8.2 & 6.2 & 19.4 \\
Yi-6B-Chat & 39.9 & 22.8 & 9.2 & 11.9 & 20.9 \\
Qwen-7B-Chat & 54.4 & 40.9 & 18.9 & 15.6 & 32.5 \\
Llama-3-8B-Instruct & 54.8 & 32.5 & 21.9 & 26.3 & 33.9 \\
InternLM2-Chat-7B & 65.1 & 51.6 & 23.4 & 19.3 & 39.8 \\
\hline
Baichuan2-13B-Chat & 40.9 & 30.2 & 12.7 & 11.3 & 23.8 \\
InternLM2-Chat-20B & 69.4 & 48.9 & 29.6 & 26.9 & 43.7 \\
Yi-34B-Chat & 69.4 & 48.9 & 25.4 & 26.9 & 42.7 \\
Qwen-14B-Chat & 70.0 & 64.3 & 33.7 & 28.7 & 49.2 \\
\hline
DeepSeek-67B-Chat & 73.7 & 50.4 & 29.7 & 35.2 & 47.3 \\
Llama-3-70B-Instruct & 69.3 & 49.3 & 36.1 & 43.1 & 49.5 \\
Qwen-72B-Chat & 82.4 & 81.0 & 50.2 & 43.8 & 64.4 \\
Qwen-1.5-110B-Chat & 90.7 & 83.9 & 57.9 & 53.7 & 71.5 \\
\hline
\multicolumn{6}{c}{\greencolor{\textit{$\triangle$Mathematical Models}}} \\
MammoTH-7B & 9.6 & 2.6 & 1.6 & 0.7 & 3.6 \\
MammoTH-13B & 24.9 & 11.5 & 7.6 & 8.7 & 13.2 \\
MetaMath-Llemma-7B & 39.0 & 25.9 & 8.9 & 8.3 & 20.5 \\
MammoTH-70B & 47.9 & 4.8 & 4.6 & 5.4 & 15.7 \\
InternLM2-Chat-Math-7B & 66.0 & 69.5 & 39.0 & 26.1 & 50.1 \\
DeepSeek-Math-7B-Instruct & 75.6 & 54.3 & 39.9 & 35.5 & 51.4 \\
DeepSeek-Math-7B-RL & 80.3 & 63.2 & 42.6 & 42.7 & 57.4 \\
InternLM2-Chat-Math-20B & 72.5 & 75.8 & 49.0 & 33.9 & 57.8 \\
\hline
\end{tabular}

        }
        \caption*{\small{Chinese Part.}}
    \end{minipage}
        \caption{\textbf{Bilingual Comparison of Models on MathBench}.}
    \label{tab: bilingual_result}
\end{table*}

The corresponding results is presented in \Cref{tab: bilingual_result}.


\section{Extra Analysis}

\newtcolorbox{AIbox}[2][]{aibox, title=#2,#1}
\begin{figure*}[!ht] 
\vspace{-5mm}
\begin{AIbox}{Example}
{
    \textbf{Question:} If $\sin x^\circ = a$, which of the following must be true for all values of $x$?
    
    \begin{itemize}
        \item[A.] $\sin (90^\circ-x^\circ) = a$
        \item[B.] $\cos (90^\circ-x^\circ) = a$
        \item[C.] $\sin (x^2)^\circ = a^2$
        \item[D.] $\cos x^\circ = a$
    \end{itemize}
    
    \textbf{Incorrect Model Answer}
    \begin{align}
        &\text{If $x = 30^\circ, \cos(90^\circ - 30^\circ) = \cos(60^\circ) = \frac{1}{2}$, but $\sin(30^\circ)  \text{not equal to } \frac{1}{2}.$}\\
        &\text{So option B is wrong ...The answer is option A.{\Huge \textcolor{red}{\ding{55}}}}
    \end{align}
    
    \textbf{Incorrect Analysis:}
    \begin{align}
        &\text{This reasoning is flawed because it misunderstands the complementary} \\ 
        &\text{angle theorem, and the model even doesn't know the basic $\sin(30^\circ) = \frac{1}{2}$.}
    \end{align}
    
    \textbf{Added with Basic Theorem}
    Given the theorem and answer the question: For a right triangle with sides $a$, $b$, and hypotenuse $c$, the correct formula for sine is $\sin = \frac{\text{opposite}}{\text{hypotenuse}}$. The complementary angle theorem states that $\sin(90^\circ - x) = \cos(x)$ and $\cos(90^\circ - x) = \sin(x)$.\{\textbf{\textit{Quesiton}}\}
    
    \textbf{Correct Analysis:}
    \begin{align}
        &\text{Given } \sin x^\circ = a, \text{ by the complementary angle theorem,} \\
        &\sin(90^\circ - x) = \cos(x). \\
        &\text{Therefore, for any value of } x, \cos(90^\circ-x^\circ) = \sin x^\circ = a. \\
        &\text{Thus, the correct answer is option B.{\Huge \textcolor{green}{\ding{51}}}}  
    \end{align}
}
\end{AIbox} 
\caption{An example of a model's limited theoretical capabilities.}
\label{fig: case of limited theory}
\vspace{-5mm}
\end{figure*}

\begin{figure*}[!ht] 
\vspace{-5mm}
\begin{AIbox}{C.2.1 Misunderstandings of Concepts}
{\color{blue}\bf Prompt:} \\
{
    \textbf{Question:} 
    Let $R$ be a ring and let $U$ and $V$ be (two-sided) ideals of $R$. Which of the following must also be ideals of $R$?   
    
    I. $\{u + v : u \in U \text{ and } v \in V\}$
    
    II. $\{uv : u \in U \text{ and } v \in V\}$
    
    III. $\{x : x \in U \text{ and } x \in V\}$\\
    \textbf{Options:} \\
    \centering
    \begin{minipage}[t]{0.8\linewidth}
    (A) II only \\
    (B) III only \\
    (C) I and II only \\
    (D) I and III only\\
    \end{minipage}
    \hfill
    \begin{minipage}[t]{0.1\linewidth}
    \flushright
    \textcolor{red}{\ding{55}} \\
    \textcolor{red}{\ding{55}} \\
    \textcolor{red}{\ding{55}} \\
    \textcolor{green}{\ding{51}} \\
    \end{minipage}
}

\textbf{Incorrect Model Answer:} \\
1. $\{u + v : u \in U \text{ and } v \in V\}$ is not necessarily an ideal of $R$. This is because the set may not be closed under the ring operation. For example, if $u \in U$ and $v \in V$, but $u + v \notin U$ or $u + v \notin V$, then $u + v$ may not be in the set $\{u + v: u \in U \text{ and } v \in V\}$.

2. $\{uv : u \in U \text{ and } v \in V\}$ is an ideal of $R$. This is because for any $r \in R$, $u \in U$, and $v \in V$, we have $r(uv) = (ru)v \in UV$, and $(uv)r = u(vr) \in UV$. Thus, $UV$ is closed under left and right multiplication by elements of $R$. Also, if $u_1v_1, u_2v_2 \in UV$, then $(u_1v_1) - (u_2v_2) = u_1v_1 - u_2v_2 = u_1v_1 + (-u_2v_2) \in UV$, since $-u_2v_2 \in UV$ because $V$ is an ideal. Therefore, $UV$ is an ideal of $R$.

3. $\{x : x \in U \text{ and } x \in V\}$ is not necessarily an ideal of $R$. This is because the set may not be closed under the ring operation. For example, if $u \in U$ and $v \in V$, but $uv \notin U$ or $uv \notin V$, then $uv$ may not be in the set $\{x: x \in U \text{ and } x \in V\}$.

Therefore, the correct answer is option II. The answer is: A\\

\textbf{Error Analysis:} \\
In crafting its response, the model fails to accurately grasp the concept of an ideal within a ring.

\end{AIbox} 
\caption{An example of misunderstandings of mathematical concepts.}
\label{fig:misunderstandings of mathematical concepts}
\end{figure*}

\begin{figure*}[!ht] 
\vspace{-5mm}
\begin{AIbox}{C.2.2 Flawed Reasoning}
{\color{blue}\bf Prompt:} \\
{
    \textbf{Question:} 
    Which of the following circles has the greatest number of points of intersection with the parabola $x^2 = y + 4$? \\
    \textbf{Options:} \\
    \centering
    \begin{minipage}[t]{0.8\linewidth}
    (A) $x^2 + y^2 = 2$ \\
    (B) $x^2 + y^2 = 9$ \\
    (C) $x^2 + y^2 = 16$ \\
    (D) $x^2 + y^2 = 1$ \\
    \end{minipage}
    \hfill
    \begin{minipage}[t]{0.1\linewidth}
    \flushright
    \textcolor{red}{\ding{55}} \\
    \textcolor{green}{\ding{51}} \\
    \textcolor{red}{\ding{55}} \\
    \textcolor{red}{\ding{55}} \\
    \end{minipage}
}
\\
\textbf{Incorrect Model Answer:} \\
\textbf{Step 1: Understand the problem}
We are given a parabola, $x^2 = y + 4$, and four circles. We need to find which circle has the maximum number of points of intersection with the parabola.

\textbf{Step 2: Analyze the options}
Let's analyze each circle one by one:

\textbf{Option A:} $x^2 + y^2 = 2$
This equation represents a circle centered at the origin $(0,0)$ with radius $\sqrt{2}$. To find the points of intersection between this circle and the parabola, we can substitute $y = x^2 - 4$ into the circle's equation and see if there are any real solutions for $x$.

\textbf{Option B:} $x^2 + y^2 = 9$
This equation represents a circle centered at the origin $(0,0)$ with radius $3$. We can follow the same process as in Option A to find the points of intersection.

\textbf{Option C:} $x^2 + y^2 = 16$
This equation represents a circle centered at the origin $(0,0)$ with radius $4$. Again, we can substitute $y = x^2 - 4$ into the circle's equation and look for real solutions.

\textbf{Option D:} $x^2 + y^2 = 1$
This equation represents a circle centered at the origin $(0,0)$ with radius $1$. We will follow the same process as before.

\textbf{Step 3: Compare the results}
After analyzing all the options, we find that:\\
Option A has two points of intersection.\\
Option B has two points of intersection.\\
Option C has four points of intersection.\\
Option D has no points of intersection.

\textbf{Step 4: Choose the best answer}
Based on our analysis, Option C has the greatest number of points of intersection with the parabola. Therefore, the correct answer is option C.\\

\textbf{Error Analysis:} \\
The model accurately grasps the mathematics of parabolas and circles, using equations to identify roots correctly. However, errors in subsequent reasoning led to inaccuracies in calculating the intersections between circles of various sizes and the specified parabola. Therefore, we classify this as a correct pathway of reasoning but with errors.

\end{AIbox} 
\caption{An example of flawed Reasoning despite logical consistency.}
\label{fig:flawed Reasoning despite logical consistency}
\end{figure*}

\begin{figure*}[!ht] 
\vspace{-5mm}
\begin{AIbox}{C.2.3 Misaligned with the Question}
{\color{blue}\bf Prompt:} \\
{
    \textbf{Question:} 
    For how many positive integers k does the ordinary decimal representation of the integer k! end in exactly 99 zeros? \\
    \textbf{Options:} \\
    \centering
    \begin{minipage}[t]{0.8\linewidth}
    (A) None \\
    (B) One \\
    (C) Four \\
    (D) Five \\
    \end{minipage}
    \hfill
    \begin{minipage}[t]{0.1\linewidth}
    \flushright
    \textcolor{red}{\ding{55}} \\
    \textcolor{red}{\ding{55}} \\
    \textcolor{red}{\ding{55}} \\
    \textcolor{green}{\ding{51}} \\
    \end{minipage}
}
\\
\textbf{Incorrect Model Answer:} \\
Let the number of trailing zeros be m.
The number is in the form of $10^m = (2^m)(5^m)$.
m must be a multiple of both 2 and 5.
There are four such numbers: 0, 10, 20, and 30.
The answer is C.\\

\textbf{Error Analysis:} \\
The model overlooks discussing the tails of the factorial of decimal numbers in the problem and instead draws an irrelevant conclusion.
\end{AIbox} 
\caption{An example of reasoning misaligned with the query’s intent.}
\label{fig:reasoning misaligned with the query’s intent}
\end{figure*}

\begin{figure*}[!ht] 
\vspace{-5mm}
\begin{AIbox}{C.2.4 Exceed max out length}
{\color{blue}\bf Prompt:} \\
{
    \textbf{Question:} 
    Let $Y_1 < Y_2 < Y_3 < Y_4 < Y_5 < Y_6$ be the order statistics of a random sample of size $n=6$ from a distribution of the continuous type having $(100 p)$th percentile $\pi_p$. Compute

    (a) $P\left(Y_2 < \pi_{0.5} < Y_5\right)$.\\
    \textbf{Options:} \\
    \centering
    \begin{minipage}[t]{0.8\linewidth}
    (A) 0.88125 \\
    (B) 0.78125 \\
    (C) 0.77625 \\
    (D) 0.7028 \\
    \end{minipage}
    \hfill
    \begin{minipage}[t]{0.1\linewidth}
    \flushright
    \textcolor{red}{\ding{55}} \\
    \textcolor{green}{\ding{51}} \\
    \textcolor{red}{\ding{55}} \\
    \textcolor{red}{\ding{55}} \\
    \end{minipage}
}
\\
\textbf{Incorrect Model Answer:} \\
    1. The order statistics $Y_1 < Y_2 < Y_3 < Y_4 < Y_5 < Y_6$ of a random sample of size $n=6$ from a distribution of the continuous type having $(100 p)$ th percentile $\pi_p$.
    
    2. We want to compute $P\left(Y_2 < \pi_{0.5} < Y_5\right)$.
    
    3. The probability $P\left(Y_2 < \pi_{0.5} < Y_5\right)$ is equal to the probability that the 2nd order statistic is less than the 50th percentile (median) and the 5th order statistic is greater than the 50th percentile (median).
    
    4. Using the concept of order statistics, we know that $Y_2$ is the 2nd smallest value in the sample, $Y_5$ is the 5th smallest value in the sample, and $\pi_{0.5}$ is the median of the distribution.
    
    5. The probability $P\left(Y_2 < \pi_{0.5} < Y_5\right)$ is equal to the probability that exactly 2 out of the 6 sample values are less than the median and exactly 5 out of the 6 sample values are greater than the median.
    
    6. This is a binomial probability problem with parameters $n=6$, $p=0.5$, and we want to find $P(X=2)$ and $P(X=5)$, where $X$ is the number of sample values less than the median.
    
    7. Using the binomial probability formula, $P(X=2) = \binom{6}{2} (0.5)^2 (0.5)^4 = 15 \times 0.25 \times 0.0625 = 0.234375$.
    
    8. Similarly, $P(X=5) = \binom{6}{5} (0.5)^5 (0.5)^1 = 6 \times 0.03125 \times 0.5 = 0.09375$.
    
    9. The probability $P\left(Y_2 < \pi_{0.5} < Y_5\right)$ is the product of these two probabilities, \\

\textbf{Error Analysis:} \\
The model demonstrates a clear grasp of quartiles and probability distributions. However, due to extensive reasoning rounds, responses often exceed the text length limit, resulting in truncation. Additionally, there is an error in the model's final statement concerning the probability product.
\end{AIbox} 
\caption{An example of responses exceeding the predefined character limit.}
\label{fig:responses exceeding the predefined character limit}
\end{figure*}

\begin{figure*}[!ht] 
\vspace{-5mm}
\begin{AIbox}{C.2.5 Responses Constrained to Options}
{\color{blue}\bf Prompt:} \\
{
    \textbf{Question:} 
    \\
    Statement 1 | If $A$ is connected, the closure of $A$ must be connected. \\
    Statement 2 | Every continuous function $f : (0, 1) \rightarrow (0, 1)$ has a fixed point.\\
    \textbf{Options:} \\
    \centering
    \begin{minipage}[t]{0.8\linewidth}
    (A) False, False\\
    (B) True, True \\
    (C) True, False \\
    (D) False, True \\
    \end{minipage}
    \hfill
    \begin{minipage}[t]{0.1\linewidth}
    \flushright
    \textcolor{red}{\ding{55}} \\
    \textcolor{red}{\ding{55}} \\
    \textcolor{green}{\ding{51}} \\
    \textcolor{red}{\ding{55}} \\
    \end{minipage}
}
\\
\textbf{Incorrect Model Answer:} \\
   The first statement is true if and only if the closure of A is connected. The second statement is false, as there are continuous functions that do not have a fixed point. 
   
   Therefore, the correct answer is option A.\\

\textbf{Error Analysis:} \\
The model directly provides an answer without elaborating on the reasoning process behind it, making it challenging to ascertain the cause of the incorrect response.
\end{AIbox} 
\caption{An example of responses constrained to options.}
\label{fig:responses constrained to options}
\end{figure*}

\begin{figure*}[!ht] 
\vspace{-5mm}
\begin{AIbox}{C.2.6 Non-adherence to the Prompt}
{\color{blue}\bf Prompt:} \\
{
    \textbf{Question:} 
    The World Series in baseball continues until either the American League team or the National League team wins four games. How many different orders are possible (e.g., ANNAAA means the American League team wins in six games) if the series goes four games?\\
    \textbf{Options:} \\
    \centering
    \begin{minipage}[t]{0.8\linewidth}
    (A) 5 \\
    (B) 4 \\
    (C) 2 \\
    (D) 6 \\
    \end{minipage}
    \hfill
    \begin{minipage}[t]{0.1\linewidth}
    \flushright
    \textcolor{red}{\ding{55}} \\
    \textcolor{red}{\ding{55}} \\
    \textcolor{green}{\ding{51}} \\
    \textcolor{red}{\ding{55}} \\
    \end{minipage}
}
\\
\textbf{Incorrect Model Answer:} \\
\textbf{Step 1}: Understand the problem The World Series in baseball continues until either the American League team or the National League team wins four games. We want to find out how many different orders are possible if the series goes exactly four games.

\textbf{Step 2}: Identify the possibilities There are two teams playing against each other, so there are only two possible outcomes for each game: either the American League team wins or the National League team wins. Since the series goes exactly four games, we can represent the outcome of each game using a single letter: \"A\" for an American League win and \"N\" for a National League win.

\textbf{Step 3}: Count the number of possible orders To count the number of possible orders, we simply need to count the number of ways we can arrange the letters \"A\" and \"N\" to form a string of length four. There are two possibilities for each position in the string, so there are a total of 2 x 2 x 2 x 2 = 16 possible orders.

\textbf{Step 4}: Check the answer choices None of the answer choices match our calculation of 16 possible orders. However, since the question asks for the number of different orders if the series goes exactly four games, and there are only two possible outcomes for each game, it stands to reason that the correct answer should be a positive integer. Therefore, the correct answer is not among the given options.
Therefore, the correct answer is not among the options A, B, C, or D.\\

\textbf{Error Analysis:} \\
The model does not provide answers in the given options.
\end{AIbox}
\caption{An example of non-adherence to the prompt.}
\label{fig:non-adherence to the prompt}
\end{figure*}

\begin{figure*}[!ht] 
\vspace{-5mm}
\begin{AIbox}{C.3.1 Reasoning Path of the Small-Scale Chat Model}
{\color{blue}\bf Prompt:} \\
{
    \textbf{Question:} 
    Alma bought a laptop computer at a store that gave a 20 percent discount off its original price. The total amount she paid to the cashier was $p$ dollars, including an 8 percent sales tax on the discounted price. Which of the following represents the original price of the computer in terms of $p$?\\
    \textbf{Options:} \\
    \centering
    \begin{minipage}[t]{0.8\linewidth}
    (A) $\frac{p}{0.88}$\\
    (B) $(0.8)(1.08) p$ \\
    (C) $\frac{p}{(0.8)(1.08)}$ \\
    (D) $0.88 p$ \\
    \end{minipage}
    \hfill
    \begin{minipage}[t]{0.1\linewidth}
    \flushright
    \textcolor{red}{\ding{55}} \\
    \textcolor{red}{\ding{55}} \\
    \textcolor{green}{\ding{51}} \\
    \textcolor{red}{\ding{55}} \\
    \end{minipage}
}
\\
\textbf{Model Answer:} \\
The original price of the laptop is $x$. After the 20\% discount, the price becomes $0.8x$. We also know that the total amount Alma paid to the cashier, including the 8\% sales tax, was $p$. So we can set up the equation:

$0.8x + 0.08x = p$

Simplifying this equation, we get:

$0.96x = p$

Now we can solve for x:

$x = p / 0.96$

$x = 0.88p$

Therefore, the original price of the computer in terms of $p$ is $0.88p$. Option D is the correct answer.

\textbf{Reasoning Path Analysis:} \\
The model's reasoning is predicated on an inaccurate interpretation of the tax calculation. Nevertheless, the mathematical reasoning employed by the model is comprehensive and logically coherent despite simple errors, indicating a sound mathematical logic within the reasoning trajectory of the small-scale chat model.
\end{AIbox} 
\caption{An example of reasoning path of the small-scale chat model.}
\label{fig:reasoning path of the small-scale chat model}
\end{figure*}

\begin{figure*}[!ht] 
\vspace{-5mm}
\begin{AIbox}{C.3.2 Reasoning Path of the Large-Scale Chat Model}
{\color{blue}\bf Prompt:} \\
{
    \textbf{Question:} 
    It is known that $\log_{2}(x)=5$. Give the value of $\log_{2}(2x)$. Explain your answer.\\
    \textbf{Options:} \\
    \centering
    \begin{minipage}[t]{0.8\linewidth}
    (A) 6\\
    (B) 12\\
    (C) 36 \\
    (D) 7 \\
    \end{minipage}
    \hfill
    \begin{minipage}[t]{0.1\linewidth}
    \flushright
    \textcolor{green}{\ding{51}} \\
    \textcolor{red}{\ding{55}} \\
    \textcolor{red}{\ding{55}} \\
    \textcolor{red}{\ding{55}} \\
    \end{minipage}
}
\\
\textbf{Model Answer:} \\
Step 1: We are given that $\log_{2}(x)=5$.

Step 2: Our goal is to find the value of $\log_{2}(2x)$.

Step 3: Let's use the property of logarithms that says $\log_b(mn) = \log_b(m) + \log_b(n)$.

Step 4: In our case, we can rewrite $2x$ as $2 \cdot x$. So, we have:

$\log_{2}(2x) = \log_{2}(2) + \log_{2}(x)$

Step 5: We know that $\log_{2}(2) = 1$, because $2^1 = 2$. And we also know that $\log_{2}(x) = 5$ (from the problem statement). So, we can substitute these values in:

$\log_{2}(2x) = 1 + 5$

Step 6: Adding 1 and 5 gives us:

$\log_{2}(2x) = 6$

Therefore, the correct answer is option A.

\textbf{Reasoning Path Analysis:} \\
The model's reasoning paths exhibit evident recursive relationships, cohesive logical connections, and well-defined objectives. Additionally, the model demonstrates adept utilization of pertinent knowledge references while responding to questions, thereby enhancing its mathematical comprehension and facilitating the provision of comprehensive and lucid reasoning paths.
\end{AIbox} 
\caption{An example of reasoning path of the large-scale chat model.}
\label{fig:reasoning path of the large-scale chat model}
\end{figure*}

\begin{figure*}[!ht] 
\vspace{-5mm}
\begin{AIbox}{C.3.3 Reasoning Path of the Math Model}
{\color{blue}\bf Prompt:} \\
{
    \textbf{Question:} 
    The first term of an arithmetic sequence is 18. The sum of the first six terms of the sequence is equal to the sum of the first seven terms.
    
    Show that the sum of the first 13 terms of the sequence is 0 and calculate the 13th term of the sequence.\\
    \textbf{Options:} \\
    \centering
    \begin{minipage}[t]{0.8\linewidth}
    (A) -18\\
    (B) -16\\
    (C) -14\\
    (D) -20\\
    \end{minipage}
    \hfill
    \begin{minipage}[t]{0.1\linewidth}
    \flushright
    \textcolor{green}{\ding{51}} \\
    \textcolor{red}{\ding{55}} \\
    \textcolor{red}{\ding{55}} \\
    \textcolor{red}{\ding{55}} \\
    \end{minipage}
}
\\
\textbf{Model Answer:} \\
13th term of the sequence is 0.

Let a be the first term and d be the common difference.

The sum of the first 6 terms is:

$S6 = 6/2[2a + (6-1)d] = 3(2a + 5d) = 6a + 15d.$

The sum of the first 7 terms is:

$S7 = 7/2[2a + (7-1)d] = 7/2(2a + 6d) = 7(a + 3d) = 7a + 21d.$

Given that $S6 = S7$, we have:

$6a + 15d = 7a + 21d.$

Rearranging terms, we get:

$a = -6d.$

The sum of the first 13 terms is:

$S13 = 13/2[2a + (13-1)d] = 13/2(2a + 12d) = 13(a + 6d).$

Substituting $a = -6d$, we get:

$S13 = 13(-6d + 6d) = 13 × 0 = 0.$

The nth term of an arithmetic sequence is given by:

$an = a + (n-1)d.$

Substituting $a = -6d$, we get:

$an = -6d + (n-1)d = d(-6 + n-1) = d(n-7).$

For the 13th term, we have:

$a13 = d(13-7) = 6d.$

But we know that $a = -6d$, so:

$a13 = 6d = -a.$

Since the first term is 18, we have $a = 18$, so:

$a13 = -18$.

Therefore, the correct answer is option A.
The answer is $\boxed{A}$. 

\textbf{Reasoning Path Analysis:} \\
The mathematical model correctly invokes the formulas and properties of the equivariant series in its reasoning path and uses mathematical reasoning to draw conclusions. It correctly applied the summation and generalization formulas of the equal difference series and used algebraic operations to derive the results. The model was derived without logical errors, each step was based on the results of previous reasoning, and the derivation was consistent with the known conditions.
\end{AIbox} 
\caption{An example of reasoning path of the large-scale chat model.}
\label{fig:reasoning path of the math model}
\end{figure*}

\label{app: prompt detail}
\subsection{Prompts Demonstration}
Please refer to the respective prompt block for a detailed demonstration.

\subsubsection{English Open-ended test}
The corresponding prompt is presented in \Cref{fig: English cloze test}.

\subsubsection{Chinese Open-ended test}
The corresponding prompt is presented in \Cref{fig: Chinese cloze test}.

\subsubsection{English single choice with reasoning}
The corresponding prompt is presented in \Cref{fig: English single choice with reasoning}.

\subsubsection{Chinese single choice with reasoning}
The corresponding prompt is presented in \Cref{fig: Chinese single choice with reasoning}.

\subsection{Error Types Demonstration}
\label{app: error analysis}
Please refer to the respective cases for a detailed error types demonstration.
\subsubsection{Misunderstandings of concepts}
The corresponding case is presented in \Cref{fig:misunderstandings of mathematical concepts}.
\subsubsection{Flawed reasoning}
The corresponding case is presented in \Cref{fig:flawed Reasoning despite logical consistency}.
\subsubsection{Misaligned with the question}
The corresponding case is presented in \Cref{fig:reasoning misaligned with the query’s intent}.
\subsubsection{Exceed max out length}
The corresponding case is presented in \Cref{fig:responses exceeding the predefined character limit}.
\subsubsection{Responses constrained to Options}
The corresponding case is presented in \Cref{fig:responses constrained to options}.
\subsubsection{Non-adherence to the prompt}
The corresponding case is presented in \Cref{fig:non-adherence to the prompt}.

\subsection{Reasoning Paths Demonstration}
\label{app: reasoning path}
\subsubsection{Small-scale chat model}
The corresponding case is presented in \Cref{fig:reasoning path of the small-scale chat model}.
\subsubsection{Large-scale chat model}
The corresponding case is presented in \Cref{fig:reasoning path of the large-scale chat model}.
\subsubsection{Math model}
The corresponding case is presented in \Cref{fig:reasoning path of the math model}.

\begin{figure*}[!ht] 
\vspace{-5mm}
\begin{AIbox}{C.1.1 English Open-ended test}  
{
{\color{blue}\bf HUMAN:}'Q: There are 15 trees in the grove. Grove workers will plant trees in the grove today. After they are done, there will be 21 trees. How many trees did the grove workers plant today?' \\
{\color{orange}\bf BOT:} 'A: We start with 15 trees. Later we have 21 trees. The difference must be the number of trees they planted. So, they must have planted 21 - 15 = 6 trees. The answer is 6.'  \\
{\color{blue}\bf HUMAN:}'Q: If there are 3 cars in the parking lot and 2 more cars arrive, how many cars are in the parking lot?'  \\
{\color{orange}\bf BOT:} 'A: There are 3 cars in the parking lot already. 2 more arrive. Now there are 3 + 2 = 5 cars. The answer is 5.'  \\
{\color{blue}\bf HUMAN:}'Q: Leah had 32 chocolates and her sister had 42. If they ate 35, how many pieces do they have left in total?'  \\
{\color{orange}\bf BOT:} "A: Leah had 32 chocolates and Leah's sister had 42. That means there were originally 32 + 42 = 74 chocolates. 35 have been eaten. So in total they still have 74 - 35 = 39 chocolates. The answer is 39."  \\
{\color{blue}\bf HUMAN:}'Q: Jason had 20 lollipops. He gave Denny some lollipops. Now Jason has 12 lollipops. How many lollipops did Jason give to Denny?'  \\
{\color{orange}\bf BOT:} 'A: Jason had 20 lollipops. Since he only has 12 now, he must have given the rest to Denny. The number of lollipops he has given to Denny must have been 20 - 12 = 8 lollipops. The answer is 8.'  \\
{\color{blue}\bf HUMAN:}'Q: Shawn has five toys. For Christmas, he got two toys each from his mom and dad. How many toys does he have now?'  \\
{\color{orange}\bf BOT:} 'A: He has 5 toys. He got 2 from mom, so after that he has 5 + 2 = 7 toys. Then he got 2 more from dad, so in total he has 7 + 2 = 9 toys. The answer is 9.'  \\
{\color{blue}\bf HUMAN:}'Q: There were nine computers in the server room. Five more computers were installed each day, from monday to thursday. How many computers are now in the server room?'  \\
{\color{orange}\bf BOT:} 'A: There are 4 days from monday to thursday. 5 computers were added each day. That means in total 4 * 5 = 20 computers were added. There were 9 computers in the beginning, so now there are 9 + 20 = 29 computers. The answer is 29.'  \\
{\color{blue}\bf HUMAN:}'Q: Michael had 58 golf balls. On tuesday, he lost 23 golf balls. On wednesday, he lost 2 more. How many golf balls did he have at the end of wednesday?'  \\
{\color{orange}\bf BOT:} 'A: Michael initially had 58 balls. He lost 23 on Tuesday, so after that he has 58 - 23 = 35 balls. On Wednesday he lost 2 more so now he has 35 - 2 = 33 balls. The answer is 33.'  \\
{\color{blue}\bf HUMAN:}'Q: Olivia has \$23. She bought five bagels for \$3 each. How much money does she have left?'  \\
{\color{orange}\bf BOT:} 'A: She bought 5 bagels for \$3 each. This means she spent 5 * \$3 = \$15 on the bagels. She had \$23 in beginning, so now she has \$23 - \$15 = \$8. The answer is 8.'  \\
{\color{blue}\bf HUMAN:}'Q: \{question\}'  \\
{\color{orange}\bf BOT:} 'A: \{answer\}'  \\
}
\end{AIbox} 
\caption{An example prompt of English Open-ended test.}
\label{fig: English cloze test}
\vspace{-5mm}
\end{figure*}

\begin{figure*}[!ht] 
\vspace{-5mm}
\begin{AIbox}{C.1.2 Chinese Open-ended test}  
{
\begin{CJK*}{UTF8}{gbsn}
{\color{blue}\bf HUMAN:} 'Q: 林中有15棵树。林务工人员今天将在林中种植树木。完成后，将有21棵树。林务工人员今天种植了多少棵树？' \\
{\color{orange}\bf BOT:} 'A: 我们从15棵树开始。后来有21棵树。差值必定是他们种植的树木数量。所以，他们必须种植了21 - 15 = 6棵树。答案是 6' \\
{\color{blue}\bf HUMAN:} 'Q: 如果停车场有3辆车，又有2辆车进来，停车场里有多少辆车？' \\
{\color{orange}\bf BOT:} 'A: 停车场已经有3辆车。又进来了2辆车。现在有3 + 2 = 5辆车。答案是 5' \\
{\color{blue}\bf HUMAN:} 'Q: 黎恩有32块巧克力，她的妹妹有42块。如果他们吃了35块，他们总共剩下多少块？' \\
{\color{orange}\bf BOT:} 'A: 黎恩有32块巧克力，Leah的妹妹有42块。这意味着原本有32 + 42 = 74块巧克力。被吃掉了35块。所以他们总共还剩下74 - 35 = 39块巧克力。答案是 39' \\
{\color{blue}\bf HUMAN:} 'Q: 杰森有20个棒棒糖。他给丹妮一些棒棒糖。现在Jason只剩下12个棒棒糖。杰森给丹妮多少个棒棒糖？' \\
{\color{orange}\bf BOT:} 'A: 杰森有20个棒棒糖。因为他现在只剩下12个，所以他必须把剩下的都给了丹妮。他给丹妮的棒棒糖数量必定是20 - 12 = 8个。答案是 8' \\
{\color{blue}\bf HUMAN:} 'Q: 莎莎有五个玩具。在圣诞节，他从他的爸爸和妈妈那里各得到了两个玩具。现在他有多少个玩具？' \\
{\color{orange}\bf BOT:} 'A: 她有5个玩具。他从妈妈那里得到了2个，所以之后他有5 + 2 = 7个玩具。然后他从爸爸那里得到了2个，所以总共他有7 + 2 = 9个玩具。答案是 9' \\
{\color{blue}\bf HUMAN:} 'Q: 服务器房间里有九台电脑。从周一到周四每天增加五台电脑。现在服务器房里有多少台电脑？' \\
{\color{orange}\bf BOT:} 'A: 从周一到周四有4天。每天增加5台电脑。这意味着总共增加了4 * 5 = 20台电脑。一开始有9台电脑，所以现在有9 + 20 = 29台电脑。答案是 29' \\
{\color{blue}\bf HUMAN:} 'Q: 迈克尔有58个高尔夫球。星期二，他丢失了23个高尔夫球。星期三，他又丢失了2个。星期三结束时他还剩下多少个高尔夫球？' \\
{\color{orange}\bf BOT:} 'A: 迈克尔一开始有58个球。星期二他丢失了23个，所以之后他还剩下58 - 23 = 35个球。星期三他又丢失了2个，所以现在他还剩下35 - 2 = 33个球。答案是 33' \\
{\color{blue}\bf HUMAN:} 'Q: 奥利弗有23美元。她用每个3美元的价格买了五个百吉饼。她还剩下多少钱？' \\
{\color{orange}\bf BOT:} 'A: 她以每个3美元的价格买了5个百吉饼。这意味着她在百吉饼上花费了5 * 3 = 15美元。她一开始有23美元，所以现在她还剩下23 - 15 = 8美元。答案是 8' \\
{\color{blue}\bf HUMAN:} 'Q: \{question\}' \\
{\color{orange}\bf BOT:} 'A: \{answer\}' \\
\end{CJK*}
}
\end{AIbox} 
\caption{An example prompt of Chinese Open-ended test.}
\label{fig: Chinese cloze test}
\vspace{-5mm}
\end{figure*}

\newpage

\begin{figure*}[!ht] 
\vspace{-5mm}
\begin{AIbox}{C.1.3 English single choice with reasoning}  
{
"Here is a multiple-choice question about mathematics. Please reason through it step by step, and at the end, provide your answer option with 'Therefore, the correct answer is option X', Where 'X' is the correct option you think from A, B, C, D. Here is the question you need to answer: \\ 
\{question\}  \\
Let's think step by step: "
}
\end{AIbox} 
\caption{An example prompt of English single choice with reasoning.}
\label{fig: English single choice with reasoning}
\vspace{-5mm}
\end{figure*}

\clearpage

\begin{figure*}[!ht] 
\vspace{-5mm}
\begin{AIbox}{C.1.4 Chinese single choice with reasoning}  
{
\begin{CJK*}{UTF8}{gbsn}
"以下是一道关于数学的单项选择题，请你一步一步推理，并在最后用“所以答案为选项X”给出答案，其中“X”为选项A，B，C，D中你认为正确的选项。下面是你要回答的问题 \\
\{question\} \\
让我们一步一步思考："
\end{CJK*}
}
\end{AIbox} 
\caption{An example prompt of Chinese single choice with reasoning.}
\label{fig: Chinese single choice with reasoning}
\vspace{-5mm}
\end{figure*}

\end{document}